\documentclass[acmsmall,screen,authorversion]{acmart}

\usepackage{enumitem}
\usepackage{subcaption}
\usepackage{multirow}
\usepackage{lscape}
\usepackage{wrapfig}
\usepackage[many]{tcolorbox}    	
\usepackage{multicol}               

\definecolor{main}{HTML}{5989cf}    
\definecolor{sub}{HTML}{cde4ff}     

\tcbset{
    sharp corners,
    colback = white,
    before skip = 0.1cm,    
    after skip = 0.1cm      
}                           

\newtcolorbox{summaryBox}{
    boxrule = 1pt, 
    leftrule = 6pt, 
    left = 1pt,
    right = 1pt,
    top = 1pt,
    bottom = 1pt
}
\AtBeginDocument{%
  }

\acmJournal{CSUR}

\begin{document}

\title{Universal Time-Series Representation Learning: A Survey}

\DeclareRobustCommand{\Fire}{\includegraphics[height=0.9em]{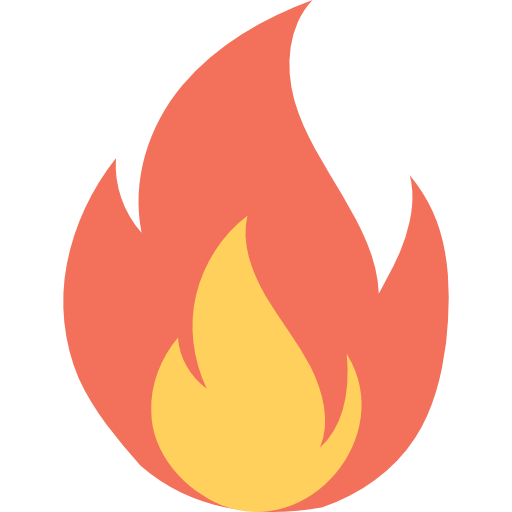}}
\DeclareRobustCommand{\Snow}{\includegraphics[height=0.9em]{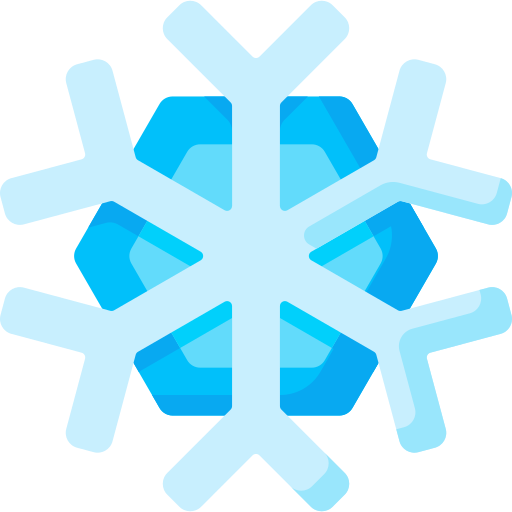}}

\author{Patara Trirat}
\email{patara.t@kaist.ac.kr}
\orcid{0000-0002-0889-813X}

\author{Yooju Shin}
\email{yooju24@gmail.com}
\orcid{0000-0002-1395-9136}

\author{Junhyeok Kang}
\email{junhyeok.kang@kaist.ac.kr}
\orcid{0009-0006-1569-7447}

\author{Youngeun Nam}
\email{youngeun.nam@kaist.ac.kr}
\orcid{0009-0008-8333-6488}

\author{Jihye Na}
\email{jihye121@kaist.ac.kr}
\orcid{0009-0001-6279-8914}

\author{Minyoung Bae}
\email{mybae@kaist.ac.kr}
\orcid{0009-0004-9267-7138}

\author{Joeun Kim}
\email{je.kim@kaist.ac.kr}
\orcid{0009-0005-2716-3904}

\author{Byunghyun Kim}
\email{rooknpown@kaist.ac.kr}
\orcid{0009-0007-5027-5343}

\author{Jae-Gil Lee}
\authornote{Jae-Gil Lee is the corresponding author.}
\email{jaegil@kaist.ac.kr}
\orcid{0000-0002-8711-7732}

\affiliation{%
  \institution{KAIST}
  \city{Daejeon}
  \country{South Korea}
}

\renewcommand{\shortauthors}{Trirat et al.}

\begin{abstract}
Time-series data exists in every corner of real-world systems and services, ranging from satellites in the sky to wearable devices on human bodies. Learning representations by extracting and inferring valuable information from these time series is crucial for understanding the complex dynamics of particular phenomena and enabling informed decisions. With the learned representations, we can perform numerous downstream analyses more effectively. Among several approaches, deep learning has demonstrated remarkable performance in extracting hidden patterns and features from time-series data without manual feature engineering. This survey first presents a novel taxonomy based on three fundamental elements in designing state-of-the-art universal representation learning methods for time series. According to the proposed taxonomy, we comprehensively review existing studies and discuss their intuitions and insights into how these methods enhance the quality of learned representations. Finally, as a guideline for future studies, we summarize commonly used experimental setups and datasets and discuss several promising research directions. An up-to-date corresponding resource is available at~\url{https://github.com/itouchz/awesome-deep-time-series-representations}.
\end{abstract}



\received{26 August 2024}
\received[revised]{31 October 2025} 
\received[accepted]{14 May 2026}

\maketitle

\section{Introduction} \label{section:introduction}

\begin{figure}[!t]
    \centering
    \includegraphics[width=\linewidth]{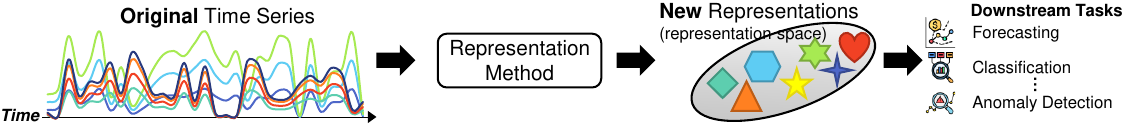}
    \caption{Basic concept of time-series representation methods.} \label{figure:tsr_concept}
    \Description{Basic concept of time-series representation methods.}
\end{figure}

A time series is a sequence of data points recorded in chronological order, reflecting the complex dynamics of particular variables or phenomena over time. Time-series data can represent various meaningful information across application domains at different time points, enabling informed decision-making and predictions, such as sensor readings in the Internet of Things\,\cite{IoT_csur, CEPWizard, trirat2023anoviz}, measurements in cyber-physical systems\,\cite{cps_attack_csur, cps_anomaly_csur}, fluctuation in stock markets\,\cite{financial_forecast_csur, ai_finance_csur}, and human activity on wearable devices\,\cite{dl_activity_reg_csur_1, dl_activity_reg_csur_2}. However, to better extract and understand meaningful information from such complicated observations, we need a \emph{mechanism to represent} these time series, which leads to the emergence of time-series representation research\,\cite{ts_survey_2012, auto_feature_ts, tsrl_non_dl_review}. Based on the new representations, we can effectively perform various downstream time-series analyses\,\cite{ts_survey_2012}, e.g., forecasting\,\cite{ts_forecasting_survey}, classification\,\cite{ts_classification_survey}, regression\,\cite{tser_survey}, and anomaly detection\,\cite{dl_ts_anomaly_review}. Fig.~\ref{figure:tsr_concept} depicts the basic concept of representation methods for time-series data. See Section~\ref{section:definitions} for formal definitions.

Early attempts\,\cite{tsrl_non_dl_review} represent time series using piecewise linear methods\,(e.g., piecewise aggregate approximation), symbolic-based methods (e.g., symbolic aggregate approximation), feature-based methods\,(e.g., shapelets), or transformation-based methods\,(e.g., discrete wavelet transform). These traditional representation methods often require considerable engineering effort and domain expertise, and may not generalize well beyond the domains or priors they were designed for\,\cite{representation_learning_tpami}. Preprocessing of real-world time-series data is also characterized as time-consuming and complex, and handcrafted feature sets sometimes fail to transfer across tasks\,\cite{tsfresh, ts_classification_survey, TS_preprocess_survey}. Since the quality of representations significantly affects the downstream task performance, many studies propose to learn the meaningful time-series representations automatically\,\cite{beyond_review_2022}. The main goal of these studies is to obtain high-quality \emph{learned} representations of time series that capture valuable information within the data and unveil the underlying dynamics of the corresponding systems or phenomena. Among several approaches, neural networks or deep learning\,(DL) have demonstrated remarkable performance in extracting hidden patterns and features from a wide range of data, including time series, without the requirement of manual feature engineering\,\cite{dtsm_survey, tsc_tser_survey, survey_dl_tsf_2025}. Note that representation learning can be categorized into task-specific\,\cite{cheng2023formertime} and \textbf{\emph{task-agnostic}}\,\cite{P88_yue2022ts2vec, P111_liu2023timesurl}, i.e., \textbf{\emph{universal}}, approaches. This survey focuses exclusively on \emph{universal} representation learning, which refers to methods that are designed for and evaluated on at least two downstream tasks. 

Given the sequential nature of time series, recurrent neural networks\,(RNNs) and their variants, such as long short-term memory and gated recurrent unit, are considered a popular choice for capturing temporal dependencies in time series\,\cite{rnn_edge_csur}. Nevertheless, RNNs are complex and computationally expensive\,\cite{survey_dl_tsf_2025}. Another line of work adopts one-dimensional convolutional neural networks\,(CNNs) to improve the computational efficiency with the parallel processing of convolutional operations\,\cite{ts_classification_nn_survey}. Although RNNs and CNNs are shown to be good at capturing temporal dependencies, they often process each variable independently and thus cannot explicitly model the relationship between different variables within a multivariate time series\,\cite{ts_gnn_survey_2023}. Many studies propose to use attention-based networks or graph neural networks to jointly learn the temporal dependencies in each variable and the correlations between different variables using attention mechanisms or graph structures\,\cite{transformer_ts_survey, ts_gnn_survey_2023}. Despite the significant progress in architectural design, time series can be collected irregularly or have missing values caused by sensor malfunctions in real-world scenarios, making the commonly used neural networks inefficient, as naive or incorrect imputation can introduce significant bias, distort the underlying temporal structure, and fail to preserve inter-variable correlations, ultimately degrading performance\,\cite{Deep_TSI_Health_Review, tsi_survey_2025}. Consequently, recent research integrates neural differential equations into existing networks such that the models can produce continuous hidden states, thereby being robust to irregular time series\,\cite{ode_ts_2019_NIPS, ts_ists_2020}.

The reliability and efficacy of DL-based methods are generally contingent upon the availability of sufficiently well-annotated data, commonly known as \emph{supervised} learning. Time-series data, however, is naturally continuous-valued, contains high levels of noise, and has less intuitively discernible visual patterns. In contrast to human-recognizable patterns in images or texts, time series can have inconsistent semantic meanings in real-world settings across application domains. As a result, obtaining a well-annotated time series is inevitably inefficient and considerably more challenging even for domain experts due to the convoluted dynamics of the time-evolving observations collected from diverse sensors or wearable devices with different frequencies. For example, we can collect a large set of sensor signals from a smart factory, while only a few of them can be annotated by domain experts. To circumvent the laborious annotation process and reduce the reliance on labeled instances, there has been a growing interest in \emph{unsupervised} and \emph{self-supervised} learning using self-generated labels from various pretext tasks without relying on human annotation\,\cite{label_efficient_tsrl_review}.

While unsupervised and self-supervised representation learning share the same objective to extract latent representations from intricate raw time series without the human-annotated labels, their underlying mechanism is different. Unsupervised learning methods\,\cite{ults_review} usually adopt autoencoders and sequence-to-sequence models to learn meaningful representations using reconstruction-based learning objectives. However, accurately reconstructing the complex time-series data remains challenging, especially with high-frequency signals. On the contrary, self-supervised learning methods\,\cite{ssl_ts_survey} leverage pretext tasks to autonomously generate labels by utilizing intrinsic information derived from the unlabeled data. Lately, pretext tasks with contrasting loss (also known as contrastive learning) have been proposed to enhance learning efficiency through discriminative pre-training with self-generated supervised signals. Contrastive learning aims to bring similar samples closer while pushing dissimilar samples apart in the feature space. These pretext tasks are self-generated challenges the model learns to solve from the unlabeled data, thereby being able to produce meaningful representations for multiple downstream tasks\,\cite{ssrl_review_2022}. 

To further enhance the representation quality and alleviate the impact of limited training samples in particular settings where collecting sufficiently large data is prohibited (e.g., human-related services), several studies also employ data-related techniques, e.g., augmentation\,\cite{P54_aboussalah2023recursive} and transformation\,\cite{P82_wu2023timesnet}, on top of the existing learning methods. Accordingly, we can effectively increase the size and improve the quality of the training data. These techniques are also deemed essential in generating pretext tasks. Different from other data types, working with time-series data needs to consider their unique properties, such as temporal dependencies and multi-scale relationships\,\cite{ts_augmentation_survey}. 

\begin{figure}[!t]
    \centering
    \includegraphics[width=\textwidth]{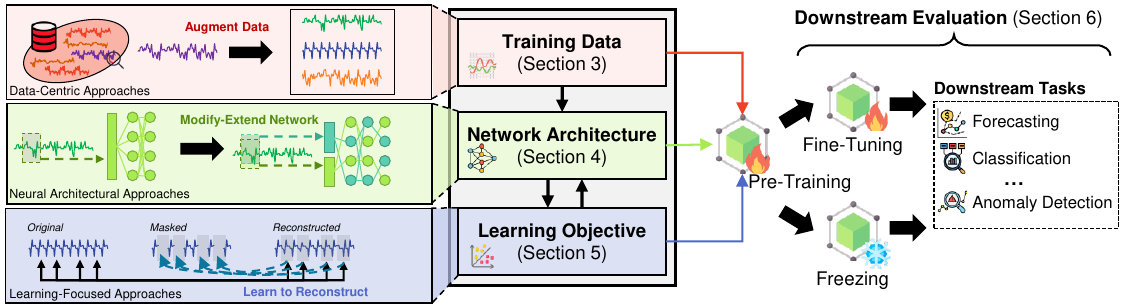}
    \Description{Key design elements and evaluation protocols of a time-series representation learning framework.}
    \caption{Key design elements and evaluation protocols of a time-series representation learning framework. \Fire{} indicates components whose parameters are actively updated in that stage, whereas \Snow{} indicates components kept frozen with no gradient updates.} \label{figure:paper_structure}
\end{figure}

\subsection{Related Surveys}

According to the background discussed above, there are three fundamental elements (also illustrated in Fig.~\ref{figure:paper_structure}) in designing a state-of-the-art universal representation learning method for time series: training data, network architectures, and learning objectives. To enhance the utility and quality of available \emph{training data}, data-related techniques (e.g., augmentation) are employed or introduced. The \emph{neural architectures} are then designed to capture underlying temporal dependencies in time series and inter-relationships between variables of multivariate time series. Last, one or multiple \emph{learning objectives} (i.e., loss functions) are newly defined to learn high-quality representations. These learning objectives are sometimes called pretext tasks if pseudo labels are generated.

Despite having the three key design elements, most existing surveys for time-series representation learning review the literature exclusively on either the neural architectural aspects\,\cite{feature_learning_ts_review, dtsm_survey} or the learning aspects\,\cite{ssl_ts_survey, ts_ptm_survey, ults_review}. An early related survey article by \citet{feature_learning_ts_review} review the DL for time series as unsupervised feature learning algorithms. The survey focuses particularly on neural architectures with a discussion on classical time-series applications. After about a decade, a few survey papers review time-series representation learning methods by focusing on learning objective aspects. For example, \citet{ssl_ts_survey} and \citet{beyond_review_2022} review self-supervised learning-based models, while, for a broader scope, \citet{ults_review} review unsupervised learning-based methods. Similarly, \citet{ts_ptm_survey} present the survey for any learning objectives with a focus on analyzing reviewed articles from transfer learning and pre-training perspectives. For a smaller scope, a survey \cite{label_efficient_tsrl_review} reviews state-of-the-art studies that specifically tackle label scarcity in time-series data, while \citet{DL_Freq_TS_Survey_KDD_25} focus exclusively on how frequency-transformation mechanisms enhance downstream task performance. With the advent of foundation and large language models\,(LLMs), recent articles\,\cite{tsfm_survey, fmtsa_survey, LLM_TS_Survey_IJCAI_24} review the adaptation of these models to time series with the main focus on learning aspects. Unlike these papers, we comprehensively review the representation learning methods for time series by focusing on their universality---effectiveness across diverse downstream tasks---with discussions on their intuitions and insights into how these methods enhance the quality of learned representations from \emph{all} three design aspects. Specifically, we aim to review and identify research directions on how recent state-of-the-art studies \emph{design the neural architecture}, \emph{devise corresponding learning objectives}, and \emph{utilize training data} to enhance the quality of the learned representations of time series for downstream tasks. \autoref{table:related_survey} summarizes the differences between our survey and the related work.

\begin{table}[!t]
\centering
\caption{Comparison of the survey scope between this article and related papers.} \label{table:related_survey}
\vspace*{-0.2cm}
\resizebox{\textwidth}{!}{%
\begin{tabular}{@{}l|cc|c|ccc|ccc@{}}
\toprule
\multirow{3}{*}{\textbf{Survey Article}} & \multicolumn{6}{c|}{\textbf{Key Design Elements in Review}} & \multicolumn{3}{c}{\textbf{Survey Coverage}} \\ \cmidrule(l){2-10} 
 & \multicolumn{2}{c|}{\textbf{Training Data}} & \multirow{2}{*}{\textbf{\begin{tabular}[c]{@{}c@{}}Neural \\ Architecture\end{tabular}}} & \multicolumn{3}{c|}{\textbf{Learning Objective}} & \multirow{2}{*}{\textbf{Universality}} & \multirow{2}{*}{\textbf{Irregularity}} & \multirow{2}{*}{\textbf{\begin{tabular}[c]{@{}c@{}}Experimental\\ Design\end{tabular}}} \\
 & \textbf{Generation} & \textbf{Augmentation} &  & \textbf{Supervised} & \textbf{Unsupervised} & \textbf{Self-Supervised} &  &  &  \\ \midrule
\citet{feature_learning_ts_review} & \textcolor{ACMRed}{$\times$} & \textcolor{ACMRed}{$\times$} & \textcolor{ACMGreen}{$\checkmark$} & \textcolor{ACMRed}{$\times$} & \textcolor{ACMRed}{$\times$} & \textcolor{ACMRed}{$\times$} & \textcolor{ACMRed}{$\times$} & \textcolor{ACMRed}{$\times$} & \textcolor{ACMRed}{$\times$} \\
\citet{beyond_review_2022} & \textcolor{ACMRed}{$\times$} & \textcolor{ACMRed}{$\times$} & \textcolor{ACMRed}{$\times$} & \textcolor{ACMRed}{$\times$} & \textcolor{ACMRed}{$\times$} & \textcolor{ACMGreen}{$\checkmark$} & \textcolor{ACMGreen}{$\checkmark$} & \textcolor{ACMRed}{$\times$} & \textcolor{ACMRed}{$\times$} \\
\citet{label_efficient_tsrl_review} & \textcolor{ACMRed}{$\times$} & \textcolor{ACMGreen}{$\checkmark$} & \textcolor{ACMRed}{$\times$} & \textcolor{ACMGreen}{$\checkmark$} & \textcolor{ACMGreen}{$\checkmark$} & \textcolor{ACMGreen}{$\checkmark$} & \textcolor{ACMGreen}{$\checkmark$} & \textcolor{ACMRed}{$\times$} & \textcolor{ACMRed}{$\times$} \\
\citet{ts_ptm_survey} & \textcolor{ACMRed}{$\times$} & \textcolor{ACMRed}{$\times$} & \textcolor{ACMRed}{$\times$} & \textcolor{ACMGreen}{$\checkmark$} & \textcolor{ACMGreen}{$\checkmark$} & \textcolor{ACMGreen}{$\checkmark$} & \textcolor{ACMGreen}{$\checkmark$} & \textcolor{ACMRed}{$\times$} & \textcolor{ACMGreen}{$\checkmark$} \\
\citet{ssl_ts_survey} & \textcolor{ACMRed}{$\times$} & \textcolor{ACMRed}{$\times$} & \textcolor{ACMRed}{$\times$} & \textcolor{ACMRed}{$\times$} & \textcolor{ACMRed}{$\times$} & \textcolor{ACMGreen}{$\checkmark$} & \textcolor{ACMRed}{$\times$} & \textcolor{ACMRed}{$\times$} & \textcolor{ACMGreen}{$\checkmark$} \\
\citet{ults_review} & \textcolor{ACMRed}{$\times$} & \textcolor{ACMRed}{$\times$} & \textcolor{ACMRed}{$\times$} & \textcolor{ACMRed}{$\times$} & \textcolor{ACMGreen}{$\checkmark$} & \textcolor{ACMGreen}{$\checkmark$} & \textcolor{ACMGreen}{$\checkmark$} & \textcolor{ACMRed}{$\times$} & \textcolor{ACMGreen}{$\checkmark$} \\
\citet{tsfm_survey} & \textcolor{ACMRed}{$\times$} & \textcolor{ACMGreen}{$\checkmark$} & Only LLMs & \textcolor{ACMGreen}{$\checkmark$} & \textcolor{ACMRed}{$\times$} & \textcolor{ACMRed}{$\times$} & \textcolor{ACMGreen}{$\checkmark$} & \textcolor{ACMRed}{$\times$} & \textcolor{ACMGreen}{$\checkmark$} \\
\citet{fmtsa_survey} & \textcolor{ACMRed}{$\times$} & \textcolor{ACMRed}{$\times$} & \textcolor{ACMGreen}{$\checkmark$} & \textcolor{ACMGreen}{$\checkmark$} & \textcolor{ACMRed}{$\times$} & \textcolor{ACMGreen}{$\checkmark$} & \textcolor{ACMRed}{$\times$} & \textcolor{ACMRed}{$\times$} & \textcolor{ACMRed}{$\times$} \\
\citet{dtsm_survey} & \textcolor{ACMRed}{$\times$} & \textcolor{ACMRed}{$\times$} & \textcolor{ACMGreen}{$\checkmark$} & \textcolor{ACMGreen}{$\checkmark$} & \textcolor{ACMRed}{$\times$} & \textcolor{ACMRed}{$\times$} & \textcolor{ACMGreen}{$\checkmark$} & \textcolor{ACMRed}{$\times$} & \textcolor{ACMGreen}{$\checkmark$} \\
\citet{LLM_TS_Survey_IJCAI_24} & \textcolor{ACMRed}{$\times$} & \textcolor{ACMRed}{$\times$} & Only LLMs & \textcolor{ACMRed}{$\times$} & \textcolor{ACMGreen}{$\checkmark$} & \textcolor{ACMGreen}{$\checkmark$} & \textcolor{ACMGreen}{$\checkmark$} & \textcolor{ACMRed}{$\times$} & \textcolor{ACMRed}{$\times$} \\
\citet{DL_Freq_TS_Survey_KDD_25} & \textcolor{ACMRed}{$\times$} & \textcolor{ACMGreen}{$\checkmark$} & \textcolor{ACMGreen}{$\checkmark$} & \textcolor{ACMRed}{$\times$} & \textcolor{ACMRed}{$\times$} & \textcolor{ACMRed}{$\times$} & \textcolor{ACMGreen}{$\checkmark$} & \textcolor{ACMRed}{$\times$} & \textcolor{ACMRed}{$\times$} \\
\midrule
This Survey & \textcolor{ACMGreen}{$\checkmark$} & \textcolor{ACMGreen}{$\checkmark$} & \textcolor{ACMGreen}{$\checkmark$} & \textcolor{ACMGreen}{$\checkmark$} & \textcolor{ACMGreen}{$\checkmark$} & \textcolor{ACMGreen}{$\checkmark$} & \textcolor{ACMGreen}{$\checkmark$} & \textcolor{ACMGreen}{$\checkmark$} & \textcolor{ACMGreen}{$\checkmark$} \\ \bottomrule
\end{tabular}%
}
\vspace*{-0.3cm}
\end{table}

\subsection{Survey Scope and Literature Collection}

For the literature review, we search for papers using the following keywords and inclusion criteria. 

\smallskip
\noindent
\textbf{Keywords}. ``time series'' AND ``representation'', ``time series'' AND ``embedding'', ``time series'' AND ``encoding'', ``time series'' AND ``modeling'', ``time series'' AND ``deep learning'', ``temporal'' AND ``representation'', ``sequential'' AND ``representation'', ``audio'' AND ``representation'', ``sequence'' AND ``representation'', and (``video'' OR ``action'') AND ``representation''. We use these keywords to search well-known repositories, including ACM Digital Library, IEEE Xplore, Google Scholar, Semantic Scholar, and DBLP, for the relevant papers. 

\smallskip
\noindent
\textbf{Inclusion Criteria}. The initial set of papers found with the above search queries are further filtered by the following criteria. Only papers meeting the criteria are included for review.

\begin{itemize}[leftmargin=9pt, noitemsep]
    \item Being written in the English language only
    \item Being a deep learning or neural networks-based approach
    \item Being published in or after 2018 at a top-tier conference or high-impact journal\footnote{Top-tier venues are evaluated based on CORE\,(\url{https://portal.core.edu.au}), KIISE\,(\url{https://www.kiise.or.kr}), or Google Scholar\,(\url{https://scholar.google.com}). Only publications from venues rated at least A by CORE/KIISE or in the top 20 in at least one subcategory by Google Scholar metrics are included for review. Recent arXiv papers also included if their authors have publication records in the qualified venues.}    
    \item Being evaluated on at least two downstream tasks using time-series, video, or audio datasets
\end{itemize}

\begin{table}[!t]
\centering
\caption{The proposed taxonomy of universal representation learning for time series. Methods may appear in multiple rows when they contribute to more than one design element.} \label{table:taxonomy}
\vspace*{-0.3cm}
\resizebox{\textwidth}{!}{%
\begin{tabular}{@{}cllll@{}}
\toprule
\textbf{Design Element} & \multicolumn{3}{c}{\textbf{Coarse-to-Fine Taxonomy}} & \multicolumn{1}{l}{\textbf{References}} \\ \midrule
\multirow{6}{*}{\begin{tabular}[c]{@{}c@{}}Data-Centric Approaches\\ (Training Data)\end{tabular}} & \multirow{3}{*}{Improving Quality ($\S$\ref{section:data_quality})} & Sample Selection &  & \cite{P01_chen2021deep, P91_franceschi2019unsupervised} \\
 &  & Time-Series Decomposition &  & \cite{P06_zeng2021contrastive, P36_behrmann2021long, P45_wang2018multilevel, P10_yang2022cross, P104_fang2023learning} \\
 &  & Input Space Transformation &  & \cite{P14_anand2021delta, P27_bilovs2022irregularly, P44_lee22multiview, P82_wu2023timesnet, P105_zhong2023multi, P106_xu2023fits} \\ \cmidrule{2-5}
 & \multirow{3}{*}{Increasing Quantity ($\S$\ref{section:data_quantity})} & \multirow{2}{*}{Augmentation} & Random Augmentation & \cite{P88_yue2022ts2vec, P57_zhang2022self, P103_zhang2023co, P111_liu2023timesurl} \\
 &  &  & Policy-Based Augmentation & \cite{P20_kim2023exploring, P22_chen2022frame, P23_kim2023frequency, P25_Zhang_2022_WACV, P54_aboussalah2023recursive, P76_luo2023time, P81_yang2022timeclr, P92_yang2022unsupervised, P98_shin23e, P99_demirel2023finding, P118_zheng2024parametric, P129_duan2024mfclr, P132_li2024unicl} \\
 &  & Generation and Curation &  & \cite{P34_nguyen2023learning, P35_zhao2023learning, P127_liu2024timer, P131_goswami2024moment} \\ 
 \midrule
\multirow{10}{*}{\begin{tabular}[c]{@{}c@{}}Neural Architectural Approaches\\ (Network Architecture)\end{tabular}} & Task-Adaptive Submodules ($\S$\ref{section:task_module}) &  &  & \cite{P01_chen2021deep, P96_WHEN_KDD23, P97_SparseTrans_KDD23, P89_liang2023units, P123_gao2024units, P125_zhang2024upme, P131_goswami2024moment} \\ \cmidrule{2-5}
 & \multirow{4}{*}{General Temporal Modeling ($\S$\ref{section:general_temporal_modeling})} & \multirow{2}{*}{Intra-Variable Modeling} & Long-Term Modeling & \cite{P03_TST_KDD, P13_tonekaboni2022decoupling, P17_zhang2023effectively, P34_nguyen2023learning, P124_donghao2024moderntcn, P38_han2020memory, P91_franceschi2019unsupervised} \\
 &  &  & Multi-Resolution Modeling & \cite{P68_liu2022tcgl, P105_zhong2023multi, P108_fraikin2023t, P34_nguyen2023learning, P02_liu2020real, P30_sanchez2019learning, P100_wang2023contrast, P72_sener2020temporal, P121_wang2024card, P126_eldele2024tslanet} \\
 &  & Inter-Variable Modeling &  & \cite{P105_zhong2023multi, P37_xie2022marina, P66_guo2021ssan, P77_cai2021time, P119_xiao2023gaformer, P46_wang2023multivariate, P121_wang2024card, P124_donghao2024moderntcn, P125_zhang2024upme, P122_wang2023fully} \\
 &  & Frequency-Aware Modeling &  & \cite{P92_yang2022unsupervised, P96_WHEN_KDD23, P82_wu2023timesnet, P106_xu2023fits, P45_wang2018multilevel, P113_zhou2023one, P126_eldele2024tslanet, P129_duan2024mfclr} \\ \cmidrule{2-5}
 & \multirow{2}{*}{Auxiliary Feature Extraction ($\S$\ref{section:aux_feature})} & Shapelet and Motif Feature Modeling &  & \cite{P07_liang2023contrastive, P115_qu2024cnn} \\
 &  & Contextual Modeling and LLM Alignment &  & \cite{P05_chen2019audio, P14_anand2021delta, P21_kim2023feat, P42_choi2023multi, P86_rahman2021tribert, P49_zhou2023one, P107_lin2023nutime, P113_zhou2023one, P83_liu2019towards, P84_li2022towards, P95_lee2022weakly, P132_li2024unicl, P50_luetto2023one, P67_chowdhury2022tarnet} \\ \cmidrule{2-5}
 & \multirow{3}{*}{\begin{tabular}[c]{@{}l@{}}Continuous Temporal and \\ Irregular Time-Series Modeling ($\S$\ref{section:cons_temporal})\end{tabular}} & Neural Differential Equations &  & \cite{P04_jhin2021attentive, P101_chen2023ContiFormer, P11_abushaqra2022crosspyramid, P19_jhin2022exit, P28_RubanovaCD19, P114_oh2024stable} \\
 &  & Implicit Neural Representations &  & \cite{P75_naour2023time, P26_fons2022hypertime} \\
 &  & Auxiliary Networks &  & \cite{P18_ma2019end, P102_zhang2023trid, P33_bianchi2019learning, P39_schirmer2022modeling, P43_shukla2021multitime, P47_ansari23a, P70_SunHSCSC021, P133_senane2024self} \\ 
 \midrule
\multirow{10}{*}{\begin{tabular}[c]{@{}c@{}}Learning-Focused Approaches\\ (Learning Objective)\end{tabular}} & Task-Adaptive Losses ($\S$\ref{section:task_loss}) &  &  & \cite{P18_ma2019end, P86_rahman2021tribert, P55_hadji2021representation, P105_zhong2023multi, P121_wang2024card, P123_gao2024units} \\ \cmidrule{2-5}
 & \multirow{3}{*}{\begin{tabular}[c]{@{}l@{}}Non-Contrasting Losses for \\ Temporal Dependency Learning ($\S$\ref{section:non_cl})\end{tabular}} & Input Reconstruction Learning &  & \cite{P05_chen2019audio, P30_sanchez2019learning, P65_nguyen2019sqn2vec, P74_li2023_TiMAE, P94_yuan2019wave2vec, P112_chen2024multi, P117_lee2024learning} \\
 &  & Masked Prediction Learning &  & \cite{P03_TST_KDD, P40_bilos23a, P63_dong2023simmtm, P67_chowdhury2022tarnet, P125_zhang2024upme, P133_senane2024self} \\
 &  & Customized Pretext Tasks &  & \cite{P08_zhang2022contrastive, P09_guo2022cross, P29_haresh2021learning, P53_wu2018random, P58_liang2022self, P60_wang2020self, P61_wang2021self, P65_nguyen2019sqn2vec, P85_duan2022transrank, P95_lee2022weakly, P104_fang2023learning, P108_fraikin2023t, P109_sun2023test, P130_dong2024timesiam, P127_liu2024timer, P128_bian2024multipatch} \\ \cmidrule{2-5}
 & \multirow{6}{*}{\begin{tabular}[c]{@{}l@{}}Contrasting Losses for \\ Consistency Modeling ($\S$\ref{section:cl})\end{tabular}} & Subseries Consistency &  & \cite{P36_behrmann2021long, P48_qian2022temporal, P60_wang2020self, P64_qian2021spatiotemporal, P87_TSRep, P91_franceschi2019unsupervised} \\
 &  & Temporal Consistency &  & \cite{P32_morgado2020learning, P122_wang2023fully, P22_chen2022frame, P41_zhang2023modeling, P29_haresh2021learning, P79_ijcai2021_324, P81_yang2022timeclr, P90_tonekaboni2021_TNC, P59_hajimoradlou2022selfsupervised, P71_yang2023tempclr} \\
 &  & Contextual Consistency &  & \cite{P88_yue2022ts2vec, P103_zhang2023co, P105_zhong2023multi, P120_xu2023retrieval, P22_chen2022frame, P98_shin23e, P80_jiao2020timeautoml, P51_chowdhury2023primenet} \\
 &  & Transformation Consistency &  & \cite{P78_jenni2021time, P79_ijcai2021_324, P56_chen2021rspnet} \\
 &  & Hierarchical and Cross-Scale Consistency &  & \cite{P07_liang2023contrastive, P34_nguyen2023learning, P12_kong2020cycle, P31_qing2022learning, P100_wang2023contrast, P25_Zhang_2022_WACV, P68_liu2022tcgl, P69_dave2022tclr, P129_duan2024mfclr} \\
 &  & Cross-Domain and Multi-Task Consistency &  & \cite{P06_zeng2021contrastive, P16_ding2022dual, P21_kim2023feat, P38_han2020memory, P42_choi2023multi, P57_zhang2022self, P110_liu2023focal, P111_liu2023timesurl, P116_lee2024soft, P132_li2024unicl} \\ 
 \bottomrule
\end{tabular}%
}
\vspace*{-0.3cm}
\end{table}

\begin{figure*}[t!]
\centering
    \begin{subfigure}[t]{0.325\linewidth}
        \includegraphics[width=\linewidth]{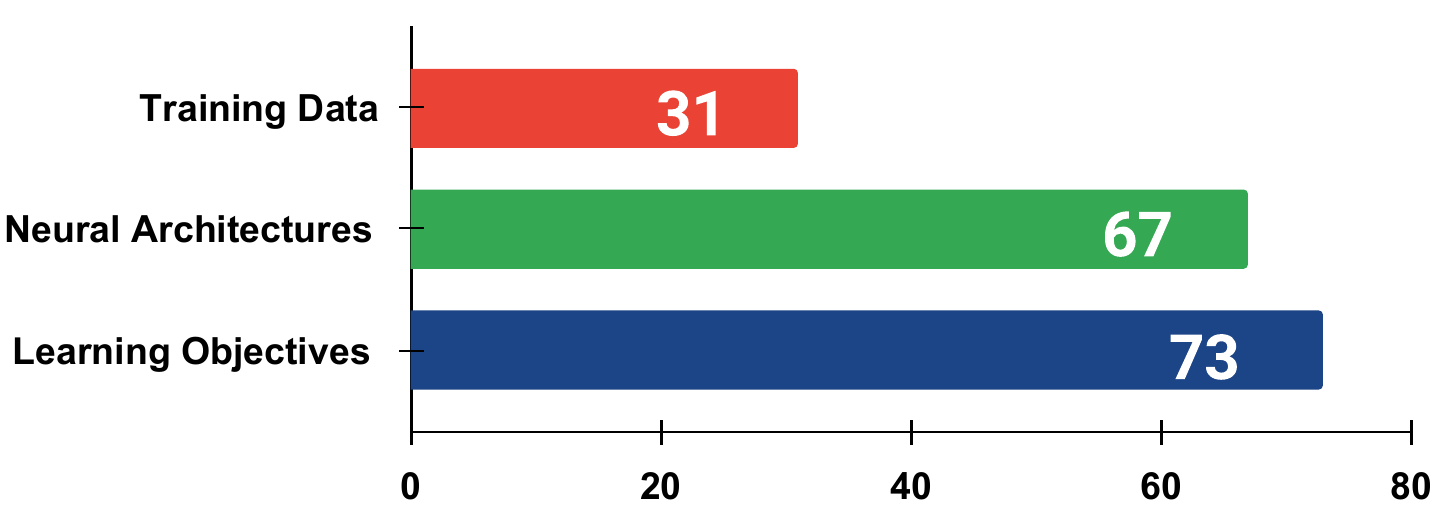}
        \vspace*{-0.5cm}   
        \caption{Focus Design Element.} \label{figure:focus_chart}
    \end{subfigure}
    \begin{subfigure}[t]{0.325\linewidth}
        \includegraphics[width=\linewidth]{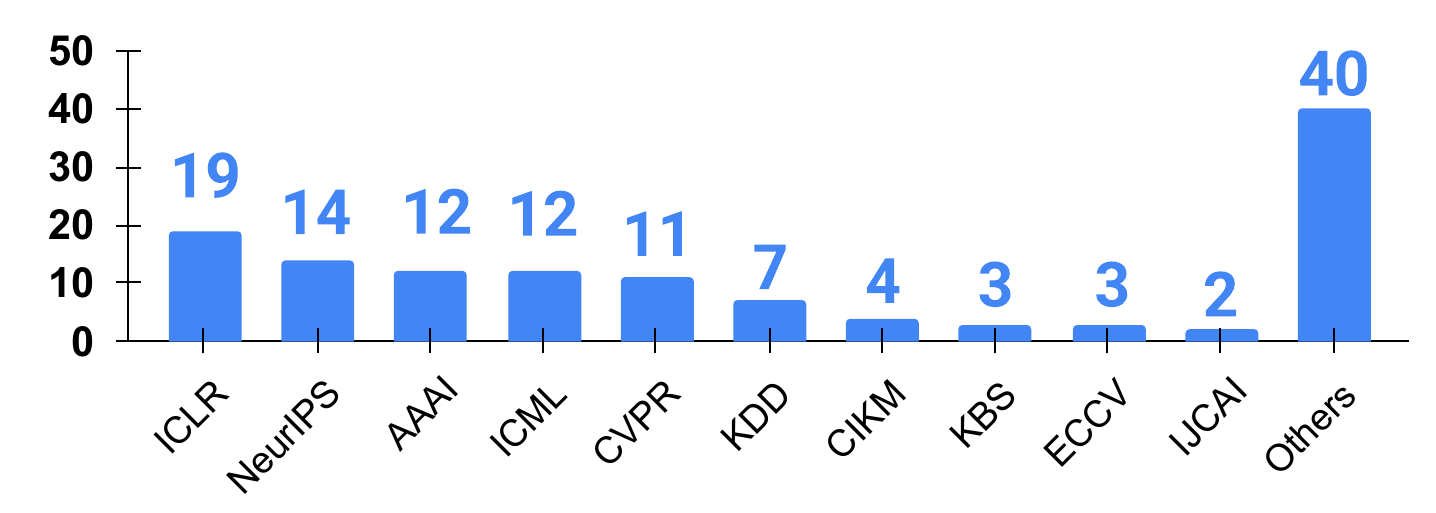}
        \vspace*{-0.5cm}   
        \caption{Publication Venue.} \label{figure:venue_chart}
    \end{subfigure}
    \begin{subfigure}[t]{0.325\linewidth}
        \includegraphics[width=\linewidth]{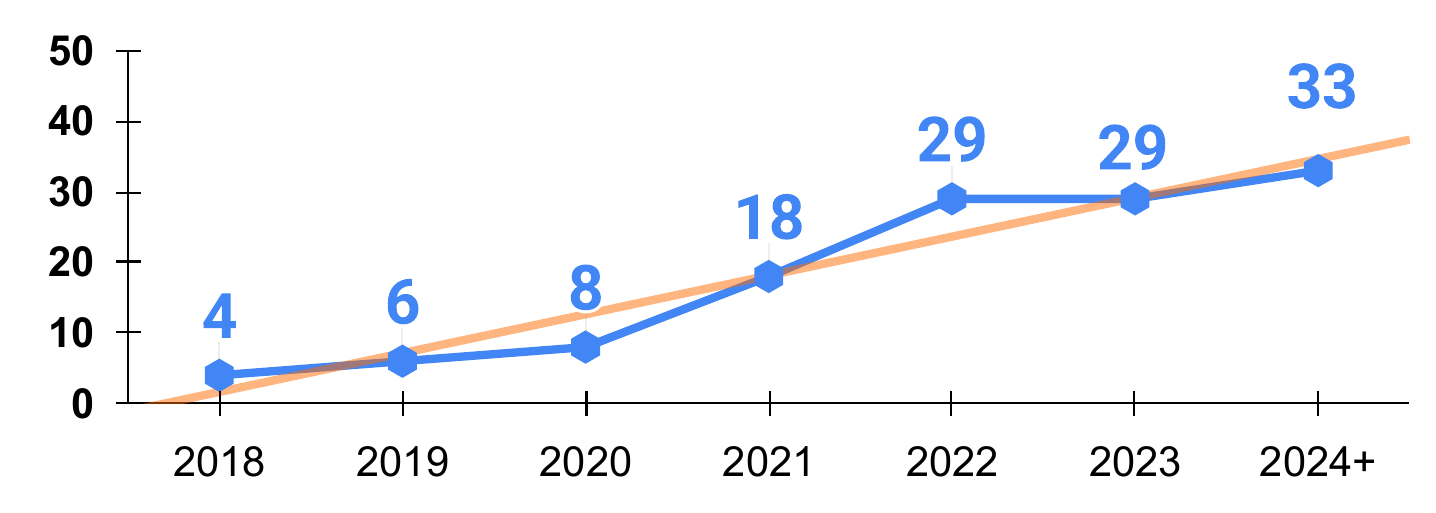}
        \vspace*{-0.5cm}   
        \caption{Publication Year.} \label{figure:year_chart}
    \end{subfigure}
\vspace*{-0.3cm}    
\caption{Quantitative summary of the selected papers.} \label{figure:quantitative_summary}
\Description{Quantitative summary of the selected papers.}
\end{figure*}

\noindent
\textbf{Quantitative Summary}. Given the above keywords and inclusion criteria, we selected \textbf{127} papers in total. Fig.~\ref{figure:quantitative_summary} shows the quantitative summary of the paper selected for review. We can notice from Fig.~\ref{figure:focus_chart} that neural architectures and learning objectives are similarly considered important in designing state-of-the-art methods. Most papers were published at ICLR, NeurIPS, followed by AAAI and ICML (Fig.~\ref{figure:venue_chart}). According to Fig.~\ref{figure:year_chart}, we expect more papers on this topic will be published in the future.

\subsection{Contributions}
This paper aims to identify what are the essential elements in designing state-of-the-art representation learning methods for time series and how these elements affect the quality of the learned representations. To the best of our knowledge, this is the first survey on universal time-series representation learning. We propose a novel taxonomy for learning universal representations of time series from the novelty perspectives---whether the main contribution of a paper focuses on what part of the design elements discussed above---to summarize the selected studies. Table~\ref{table:taxonomy} outlines and compares the reviewed articles based on the proposed taxonomy. We begin by exploring the essence of data-driven methods, categorizing data-centric approaches from the papers that primarily aim to enhance the usefulness of the training data. From the perspective of network architectures, we review the evolution of neural architectures by focusing on how these advancements enhance the quality of learned representations. From the learning perspective, we classify how different objectives enhance the generalizability of learned representations across diverse downstream tasks. It is important to note that these three categories are not mutually exclusive; they represent a design space. A single state-of-the-art method often incorporates novel contributions across multiple categories. Therefore, a given paper may be discussed in several sections of this survey, each time from the perspective of its contribution to that specific design element. Overall, our main contributions are as follows.
\begin{itemize}[leftmargin=9pt, noitemsep]
    \item We conduct an extensive literature review of universal time-series representation learning based on a novel and up-to-date taxonomy that categorizes the reviewed methods into three main categories: data-centric, neural architectural, and learning-focused approaches.
    \item We provide a guideline on the experimental setup and benchmark datasets for assessing representation learning methods for time series.
    \item We discuss several open research challenges and new insights to facilitate future work.
\end{itemize}

\noindent
\textbf{Article Organization.} Section~\ref{section:prelims} introduces the definitions and specific background knowledge regarding time-series representation learning. Section~\ref{section:data_centric} gives a review on data-centric approaches. In Section~\ref{section:architectures}, we present an extensive review of the methods focusing on neural architecture aspects. We then discuss the methods focusing on deriving new learning objectives in Section~\ref{section:learning_method}. In addition, we discuss the evaluation protocol for time-series representation learning and promising future research directions in Section~\ref{section:evaluation} and Section~\ref{section:challenges_and_trends}, respectively. Finally, Section~\ref{section:conclusion} concludes this survey.

\section{Preliminaries} \label{section:prelims}
This section presents definitions and notations used throughout this paper, descriptions of downstream time-series analysis tasks, and unique properties of time series.

\begin{figure}[H]
    \centering
    \vspace*{-0.3cm}
    \includegraphics[width=0.75\linewidth]{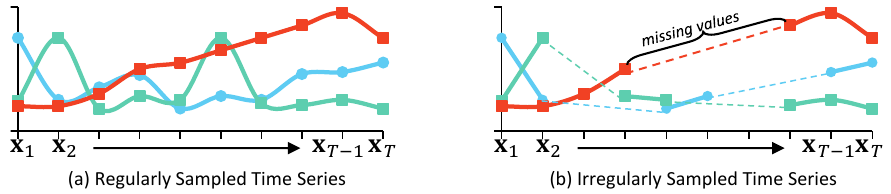}
    \vspace*{-0.3cm}
    \caption{Illustrations of regularly and irregularly sampled multivariate time series ($V = 3$).} \label{figure:compare_irregular}
    \Description{Illustrations of regularly and irregularly sampled multivariate time series with V = 3.}
    \vspace*{-0.3cm}
\end{figure}

\subsection{Definitions} \label{section:definitions}
\begin{definition}[Time Series] \label{def:time_series}
A time series $\mathbf{X}$ is a chronologically ordered sequence of $V$-variate data points recorded at specific time intervals. $\mathbf{X} = (\mathbf{x}_1, \dots, \mathbf{x}_t, \mathbf{x}_T)$, where $\mathbf{x}_t \in \mathbb{R}^V$ is the observed value at $t$-th time step, $V$ is the number of variables (also known as channels), and $T$ is the length of the time series. When $V = 1$, it becomes the univariate time series; otherwise, it is multivariate time series. Audio and video data can be considered special cases of time series with more dimensions. The time intervals are typically equally spaced, and the values can represent any measurable quantity, such as temperature, sales figures, or any phenomenon that changes over time. 
\end{definition}

\begin{definition}[Irregularly-Sampled Time Series] \label{def:irregular_time_series}
An irregularly-sampled time series\,(ISTS) is a time series where the intervals between observations are \textbf{\emph{not}} consistent or regularly spaced. Thus, the time intervals between $(\mathbf{x}_1, \mathbf{x}_2)$ and $(\mathbf{x}_2, \mathbf{x}_3)$ is unequal, as illustrated in Fig.~\ref{figure:compare_irregular}. It is often encountered in situations where data is collected opportunistically or when events occur irregularly and sporadically, e.g., sensor malfunctions, leading to varying time gaps between observations.
\end{definition}


\begin{definition}[Time-Series Representation Learning] \label{def:tsrl}
Given a raw time series $\mathbf{X}$, time-series representation learning aims to learn an encoder $f_{e}$, an embedding\,(often nonlinear) function that maps $\mathbf{X}$ into a $R$-dimensional representation vector $\mathbf{Z} = (\mathbf{z}_1, \dots, \mathbf{z}_{R-1}, \mathbf{z}_R)$ in the latent space, where $\mathbf{z}_i \in \mathbb{R}^{F}$. $\mathbf{Z}$ usually has either equal ($R = T$) or shorter ($R < T$) length of the original time series. When $R = T$, $\mathbf{Z}$ is timestamp-wise (or point-wise) representation that contains representation vectors $\mathbf{z}_i$ with feature size $F$ for each $t$. In contrast, when $R < T$, $\mathbf{Z}$ is the compressed version of $\mathbf{X}$ with reduced dimension, and $F$ is usually $1$, producing the series- or instance-wise representation.
\end{definition}

A crucial measure to assess the quality of a representation learning method, i.e., the encoder $f_{e}$, is its ability to produce the representations $\mathbf{Z}$ that effectively facilitate downstream tasks, either with or without fine-tuning (see Section~\ref{section:evaluation}). Once we obtain $\mathbf{Z}$, we can use it as input for downstream tasks to evaluate its actual performance. Here, we define the common downstream tasks as follows.

\begin{definition}[Forecasting] \label{def:tsf}
Time-series forecasting\,(TSF) aims to predict the future values of a time series. This is typically achieved by explicitly modeling the dynamics and dependencies among historical observations. It can be short-term or long-term forecasting depending on the prediction horizon $H$. Formally, given a time series $\mathbf{X}$, TSF predicts the next $H$ values $(\mathbf{x}_{T+1}, \dots, \mathbf{x}_{T+H})$ that are most likely to occur. 
\end{definition}

\begin{definition}[Classification] \label{def:tsc}
Time-series classification\,(TSC) aims to assign predefined class labels $\mathbf{C} = \{c_1, \dots, c_{|\mathbf{C}|}\}$ to a set of time series. Let $\mathcal{D} = \{(\mathbf{X}_i, \mathbf{y}_i)\}_{i = 1}^N$ denote a time-series dataset with $N$ samples, where $\mathbf{X}_i \in \mathbb{R}^{T \times V}$ is a time series and $\mathbf{y}_i \in \mathbf{C}$ is the corresponding class label from a set of predefined labels $\mathbf{C}$. Formally, TSC aims to learn a function that maps a given time-series instance $\mathbf{X}_i$ to its correct class label $\mathbf{y}_i$. 
\end{definition}

\begin{definition}[Extrinsic Regression] \label{def:tser}
Time-series extrinsic regression\,(TSER) shares a similar goal to TSC with a key difference in label annotation. While TSC predicts a categorical value, TSER predicts a continuous value for a variable external to the input time series. That is, $y_i \in \mathbb{R}$. Formally, TSER trains a regression model to map a given time series $\mathbf{X}_i$ to a numerical value $y_i$. In this context, forecasting with $H = 1$ (Definition~\ref{def:tsf}) can be considered a form of intrinsic regression, as it predicts future values of the input variables themselves.
\end{definition}

\begin{definition}[Clustering] \label{def:tscl}
Time-series clustering\,(TSCL) is the process of finding natural groups, called clusters, in a set of time series $\mathcal{X} = \{\mathbf{X}_i\}_{i = 1}^N$. It aims to partition $\mathcal{X}$ into a group of clusters $\mathbf{G} = \{g_1, \dots, g_i, g_{|\mathbf{G}|}\}$ by maximizing the similarities between time series within the same cluster and the dissimilarities between time series of different clusters. Formally, given a similarity measure $f_{s}(\cdot,\cdot)$, $\forall i_1,i_2,j~f_{s}(\mathbf{X}_{i_1}, \mathbf{X}_{j}) \gg f_{s}(\mathbf{X}_{i_1}, \mathbf{X}_{i_2})~\text{for}~\mathbf{X}_{i_1}, \mathbf{X}_{i_2} \in g_i$ and $\mathbf{X}_{j} \in g_j$. 
\end{definition}

\begin{definition}[Segmentation] \label{def:tss}
Time-series segmentation\,(TSS) aims to assign a label to a subsequence $\mathbf{X}_{T_s,T_e}$ of $\mathbf{X}$, where $T_s$ is the start offset and $T_e$ is the end offset, consisting of contiguous observations of $\mathbf{X}$ from time step $T_s$ to $T_e$. That is, $\mathbf{X}_{T_s,T_e} = (\mathbf{x}_{T_s}, \dots, \mathbf{x}_{T_e})$ and $1 \le T_s \le T_e \le T$. Let a change point\,(CP) be an offset $i \in [i, \dots, T]$ w.r.t. to a state transition in time series, TSS finds a set of segmentation of $\mathbf{X}$, having the ordered sequence of CPs in $\mathbf{X}$ (i.e., $\mathbf{x}_{i_1}, \dots, \mathbf{x}_{i_S}$) with $1 < i_1 < \cdots < i_S < T$ at which the state of observations changed. After identifying the number and locations of all CPs, we can set the start offset $T_s$ and end offset $T_e$ for each segment in $\mathbf{X}$.
\end{definition}

\begin{definition}[Anomaly Detection] \label{def:tsad}
Time-series anomaly detection\,(TSAD) aims to identify abnormal time points that significantly deviate from the other observations in a time series. Commonly, TSAD learns the representations of normal behavior from a time series $\mathbf{X}$. Then, the trained model computes anomaly scores $\mathbf{A} = (a_i, \dots, a_{|\mathbf{X}^{\prime}|})$ of all values in an unseen time series $\mathbf{X}^{\prime}$ to determine which time point in $\mathbf{X}^{\prime}$ is anomalous. The final decisions are obtained by comparing each $a_i$ with a predefined threshold $\delta$: anomalous if $a_i > \delta$ and normal otherwise.
\end{definition}

\begin{definition}[Imputation of Missing Values] \label{def:tsi}
Time-series imputation\,(TSI) aims to fill missing values in a time series with realistic values to facilitate subsequent analysis. Given a time series $\mathbf{X}$ and a binary $M = (m_1, \dots, m_t, m_T)$, $\mathbf{x}_t$ is missing if $m_t = 0$, and is observed otherwise. Let $\mathbf{\hat{X}}$ denote the predicted values generated by a TSI method, the imputed time series $\mathbf{X}_{\text{imputed}} = \mathbf{X} \odot M + \mathbf{\hat{X}} \odot (1 - M)$.
\end{definition}

\begin{definition}[Retrieval] \label{def:tsr}
Time-series retrieval\,(TSR) aims to obtain a set of time series that are most similar to a query provided. Given a query time series $\mathbf{X}_q$ and a similarity measure $f_{s}(\cdot, \cdot)$, find an ordered list $\mathcal{Q} = \{\mathbf{X}_i\}_{i = 1}^{K}$ of time series in the given dataset or database, containing $K$ time series that are the most similar to $\mathbf{X}_q$ according to $f_{s}$.
\end{definition}

Following Definition~\ref{def:tsrl}, we can use the corresponding representation $\mathbf{Z} = f_{e}(\mathbf{X})$ to perform the above downstream tasks instead of the raw time series $\mathbf{X}$.

\subsection{Unique Properties of Time Series}
In this subsection, we discuss unique properties in time-series data that have been explored by existing studies for time-series representation learning\,\cite{feature_learning_ts_review, ts_ptm_survey}. Due to the following properties,  techniques devised for image or text data are usually difficult to transfer directly to time series.

\subsubsection{Temporal Dependency} Time series exhibits dependencies on the time variable, where a data point at a given time correlates with its previous values. Given an input $\mathbf{x}_t$ at time $t$, the model predicts $y_t$, but the same input at a later time could be a different prediction. Therefore, windows or subsequences of past observations are usually included as inputs to the model to learn such temporal dependency. The length of windows for capturing the time dependencies could also be unknown. There are local and global temporal dependencies. The former is usually associated with abrupt changes or noises, while the latter is associated with collective trends or recurrent patterns.

\subsubsection{High Noise and Dimension} Time-series data, especially in real-world environments, often contain noises and have high dimensions. These noises can arise from sources such as measurement errors from faulty sensors, human mistakes, or other environmental uncertainties\,\cite{Noise_GPS_TS, tsad_survey_csur}. Dimensionality reduction techniques and wavelet transforms can address this issue by filtering some noises and reducing the dimension of the raw time series\,\cite{tsrl_non_dl_review}. Nevertheless, we may lose valuable information and need domain-specific knowledge to select the suitable dimensionality reduction and filtering techniques.

\subsubsection{Inter-Relationship across Variables} This characteristic occurs exclusively in multivariate time series. It is often uncertain whether there is sufficient information to understand a phenomenon of a time series when only a limited number of variables are analyzed as there may exist relationships between variables underlying the process or state. For example, in traffic forecasting, variables such as traffic flow, speed, and occupancy are highly correlated. A model that explicitly captures these inter-variable relationships is better equipped to make accurate predictions\,\cite{ts_gnn_survey_2023}. Similarly, in financial data, monitoring a single stock representing a fraction of a complex system may not provide enough information to forecast future values\,\cite{ai_finance_csur}.

\subsubsection{Variability and Nonstationarity} Time-series data also possess variability and nonstationarity properties, meaning that statistical characteristics, such as mean, variance, and frequency, change over time. These changes usually reveal seasonal patterns, trends, and fluctuations. Here, seasonality refers to repeating patterns that regularly appear, while trends describe long-term changes or shifts over time. In some cases, the change in frequency is so relevant to the task that it is more beneficial to work in the frequency domain than in the time domain. 

\subsubsection{Diverse Semantics} In contrast to image and text data, learning universal representations of time series is challenging due to the lack of a large-scale unified semantic time-series dataset. For instance, each word in a text dataset has similar semantics in different sentences with high probability. Accordingly, the word embeddings learned by the model can transfer across different scenarios. However, time-series datasets are challenging to obtain subsequences (corresponding to words in text sequences) that have consistent semantics across scenarios and applications, making it difficult to transfer the knowledge learned by the model. This property also makes time-series annotation tricky and challenging, even for domain experts.

\subsection{Neural Architectures for Time Series} \label{section:basic_nn} 
Choosing an appropriate neural architecture to model complex temporal and inter-relationship dependencies in time series is an essential part of the design elements. This subsection overviews basic neural network architectures used as building blocks in state-of-the-art representation learning methods for time series.

\subsubsection{Multi-Layer Perceptrons (MLP)}
The most basic neural network architecture is the MLP\,\cite{ts_classification_survey}, i.e., fully connected\,(FC) network. In MLP models, the number of layers and neurons (or hidden units) are hyperparameters to be defined. Specifically, all neurons in a layer are connected to all neurons of the following layer. These connections contain weights in the neural network. The weights are later updated by applying a non-linearity to an input. As MLP-based models process input data in a single fixed-length representation without considering the temporal relationships between the data points, they are basically unsuitable for capturing the temporal dependencies and time-invariant features. Each time step is weighted individually, and time-series elements are learned independently from each other. 
 

\subsubsection{Recurrent Neural Networks (RNN)} RNN\,\cite{rnn_physiology} is a neural architecture with internal memory (i.e., state) specifically designed to process sequential data, thus suitable for learning temporal features in time series. The memory component enables RNN-based models to refer to past observations when processing the current one, resulting in improved learning capability. RNNs can also process variable-length inputs and produce variable-length outputs. This capability is enabled by sharing parameters over time through direct connections between layers. However, they are ineffective in modeling long-term dependencies, besides being computationally expensive due to sequential processing. RNN-based models are usually trained iteratively using a technique called back-propagation through time. Unfolded RNNs are similar to deep networks with shared parameters connected to each RNN cell. Due to the depth and weight-sharing in RNNs, the gradients are summed up at each time step to train the model, undergoing continuous matrix multiplication due to the chain rule. Thus, the gradients often either shrink exponentially to small values (i.e., vanishing gradients) or blow up to large values (i.e., exploding gradients). These problems give rise to long short-term memory\,(LSTM) and gated recurrent unit\,(GRU).


\paragraph{Long Short-Term Memory (LSTM)} LSTM\,\cite{LSTM} addresses the well-known vanishing and exploding gradient problems in the standard RNNs by integrating memory cells with a gating mechanism (i.e., cell state gate, input gate, output gate, and forget gate) into their state dynamics to control the information flow between cells. As the inherent nature of LSTM is the same as RNN, LSTM-based models are also suitable for learning sequence data like time series and video representation learning, with a better capability to capture long-term dependencies.

\paragraph{Gated Recurrent Unit (GRU)} GRU\,\cite{GRU} is another popular RNN variant that can control information flow and memorize states across multiple time steps, similar to LSTM, but with a simpler cell architecture. Compared to LSTM, GRU cells have only two gates (reset and update gates), making it more computationally efficient and requiring less data to generalize. 

\subsubsection{Convolutional Neural Networks (CNN)} CNN\,\cite{ts_classification_survey} is a very successful neural architecture, proven in many domains, including computer vision, speech recognition, and natural language processing. Accordingly, researchers also adopt CNN for time series. To use CNN for time series, we need to encode the input data into an image-like format. The CNN receives embedding of the value at each time step and then aggregates local information from nearby time steps using convolution layers. The convolution layer, consisting of several convolution kernels (i.e., filters), aims to learn feature representations of the inputs by computing different feature maps. Each neuron of a feature map connects to a region of neighboring neurons in the previous layer called the receptive field. The feature maps can be created by convolving the inputs with learned kernels and applying an element-wise nonlinear activation to the convolved results. Here, all spatial locations of the input share the kernel for each feature map, and several kernels are used to obtain the entire feature map. Many improvements have been made to CNN, such as using deeper networks, applying smaller and more efficient convolutional filters, adding pooling layers to reduce the resolution of the feature maps, and utilizing batch normalization to improve the training stability. As standard CNNs are designed for processing images, widely used CNN architectures for time series are one-dimensional CNN and temporal convolutional networks\,\cite{label_efficient_tsrl_review}.  


\paragraph{Temporal Convolutional Networks (TCN)} Different from the standard CNNs, TCN\,\cite{TCN} uses the fully convolutional network to make all layers the same length and employ casual convolution operation to avoid information leakage from the future time step to the past. Compared to RNN-based models, TCN has recently shown to be more accurate, simpler, and more efficient across diverse downstream tasks\,\cite{ts_ptm_survey}.

\subsubsection{Graph Neural Networks (GNN)} GNN\,\cite{ts_gnn_survey_2023} aims to learn directly from graph representations of data. A graph consists of nodes and edges, with each edge connecting two nodes. Both nodes and edges can have associated features. Edges can be directional or un-directional and can be weighted. Graphs better represent data not easily represented in Euclidean space, such as spatio-temporal data like the electroencephalogram and traffic monitoring networks. GNNs receive the graph structure and any associated node and edge attributes as input. Typically, the core operation in GNNs is graph convolution, which involves exchanging information across neighboring nodes. This operation enables the GNN-based model to explicitly rely on the inter-variable dependencies represented by the graph edges for processing multivariate time-series data. While both RNNs and CNNs perform well on Euclidean data, time series are often more naturally represented as graphs in many scenarios. Consider a network of traffic sensors where the sensors are not uniformly spaced, deviating from a regular grid. Representing this data as a graph captures its irregularity more precisely than a Euclidean space. However, using standard deep learning algorithms to learn from graph structures is challenging as nodes may have varying numbers of neighboring nodes, making it difficult to apply the convolution operation. Thus, GNNs are more suitable to graph data.

\subsubsection{Attention-based Networks} The attention mechanism was introduced by \citet{attention} to improve the performance of encoder-decoder models in machine translation. The encoder encodes a source sentence into a vector in latent space, and the decoder then decodes the latent vector into a target language sentence. The attention mechanism enables the decoder to pay attention to the segments of the source for each target through a context vector. Attention-based neural networks are designed to capture long-range dependencies with broader receptive fields, usually lacking in CNNs and RNNs. Thus, attention-based models provide more contextual information to enhance the models’ learning and representation capability. The underlying mechanism (i.e., attention mechanism) is proposed to make the model focus on essential features in the input while suppressing the unnecessary ones. For instance, it can be used to enhance LSTM performance in many applications by assigning attention scores for LSTM hidden states to determine the importance of each state in the final prediction. Moreover, the attention mechanism can improve the interpretability of the model. However, it can be more computationally expensive due to the large number of parameters, making it prone to overfitting when the training data is limited.

\paragraph{Self-Attention Module} Self-attention has been demonstrated to be effective in various natural language processing tasks due to its ability to capture long-term dependencies in text\,\cite{tsc_tser_survey}. The self-attention module is usually embedded in encoder-decoder models to improve the model performance and leveraged in many studies to replace the RNN-based models to improve learning efficiency due to its fast parallel processing.


\paragraph{Transformers} The unprecedented performance of stacking multi-headed attention, called Transformers\,\cite{Transformers}, has led to numerous endeavors to adapt multi-headed attention to time-series data. Transformers for time series usually contain a simple encoder structure consisting of multi-headed self-attention and feed-forward layers. They integrate information from data points in the time series by dynamically computing the associations between representations with self-attention. Thanks to the practical capability to model long-range dependencies, Transformers have shown remarkable performance in sequence data. Many recent studies use Transformers as the backbone architecture for time-series analysis.\,\cite{transformer_ts_survey}.

\subsubsection{Neural Ordinary Differential Equations (Neural ODE)} Let $f_{\theta}$ be a function specifying continuous dynamics of a hidden state $\mathbf{h}(t)$ with paramters $\theta$. Neural ODEs are one of the continuous-time models that define the hidden state $\mathbf{h}(t)$ as a solution to ODE initial-value problem $\frac{d\mathbf{h}(t)}{dt} = f_{\theta}(\mathbf{h}(t), t)$ where $\mathbf{h}(t_o) = \mathbf{h}_0$. The hidden state $\mathbf{h}(t)$ is defined at all time steps, and can be evaluated at any desired time steps using a numerical ODE solver. Formally, $\mathbf{h}_0, \dots, \mathbf{h}_T = \text{ODESolver}(f_{\theta}, \mathbf{h}_0, (t_0, \dots, t_T))$. For training ODE-based deep learning models using black-box ODE solvers, we can use the adjoint sensitivity method to compute memory-efficient gradients w.r.t. the neural network parameters $\theta$, as described by \citet{ode_ts_2019_NIPS}. Neural ODEs are usually combined with RNN or its variants to sequentially update the hidden state at observation times\,\cite{ts_ists_ode_2020}. These models provide an alternative recurrence-based solution with better properties than traditional RNNs in terms of their ability to handle irregularly sampled time series. 
\section{Data-Centric Approaches} \label{section:data_centric}
\begin{figure}[!t]
    \centering
    \includegraphics[width=\linewidth]{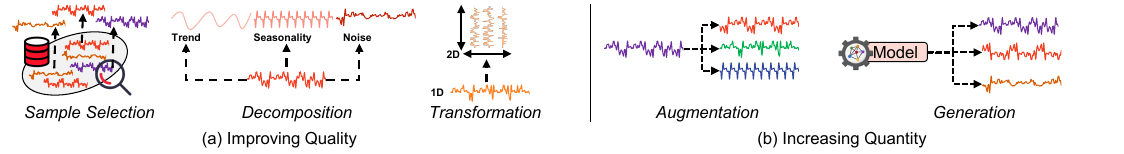}
    \vspace*{-0.7cm}
    \caption{Taxonomic gallery of data-centric approaches. This figure expands only the training-data block of the overall framework in Fig.~\ref{figure:paper_structure} by illustrating methods that either improve data quality or increase data quantity.} \label{figure:data_centric}
    \Description{Illustrative examples of data-centric approaches. (a) Improving quality focuses on selecting the most useful samples or extracting underlying properties from available training data, while (b) increasing data quantity aim to increase the size and diversity of the data.}
    \vspace*{-0.5cm}
\end{figure}

As shown in Fig.~\ref{figure:paper_structure}, training data is the first design element in universal time-series representation learning. This section zooms into that element and reviews methods that enhance the usefulness of the \emph{available training data} rather than primarily modifying the model architecture or the loss function. Fig.~\ref{figure:data_centric} provides the section-level taxonomy: data-centric approaches either improve data quality or increase data quantity.

\subsection{Improving Data Quality} \label{section:data_quality}

\subsubsection{Sample Selection} \label{section:sample_selection} Sample selection aims to effectively choose the best samples from available training data for a particular learning scenario. An early study adopting sample selection for time-series representation learning is proposed by \citet{P91_franceschi2019unsupervised}. This work uses time-based negative sampling which determines several negative samples by independently choosing sub-series from other time series at random, whereas a sub-series within the referenced time series is consider a positive sample. This technique encourages representations of the input time segment and its sampled sub-series to be close to each other. MTRL\,\cite{P01_chen2021deep} selects discriminative pairs using a distance-weighted sampling strategy that, based on the pairwise dynamic time warping\,(DTW) distance matrix, assigns higher sampling probabilities to more similar pairs\,(i.e., those with smaller DTW distances) after converting distances into a probability matrix. This approach helps achieve faster convergence and higher accuracy. Here, the term \emph{discriminative} refers to negatives yielding a higher gradient signal\,(e.g., informative ones) and contributing more to representation shaping.

\subsubsection{Time-Series Decomposition}  mWDN\,\cite{P45_wang2018multilevel} is an early attempt that integrates multi-level discrete wavelet decomposition into a deep neural network framework. This framework allows for the preservation of frequency learning advantages while facilitating the fine-tuning of parameters within a deep learning context. \citet{P104_fang2023learning} decompose the spatial relation within multivariate time series into prior and dynamic graphs to model both common relation shared across all instances and distinct correlation that varies across instances.

To enhance video-level tasks, including video classification, and more detailed tasks like action segmentation, \citet{P36_behrmann2021long} separate the representation space into stationary and non-stationary characteristics through contrastive learning from long and short views. \citet{P06_zeng2021contrastive} improve generalization across various downstream tasks by learning spatially-local/temporally-global and spatially-global/temporally-local features from audio-visual modalities to capture global and local information inside a video. This method enables the capturing of both slowly changing patch-level information and fast changing frame-level information. More recently, \citet{P10_yang2022cross} propose a unified framework for joint audio-visual speech recognition and synthesis. Each modality is decomposed into modality-specific and modality-invariant features. Modality-invariant codebook enhances the alignment between the linguistic representation space of visual and audio modalities.

\subsubsection{Input Space Transformation} To effectively exploit vision models, DeLTa\,\cite{P14_anand2021delta} transforms 1D time series into 2D images and use the transformed images as features for the subsequent learning phase with models pretrained on large image datasets, such as ResNet50. Similarly, TimesNet\,\cite{P82_wu2023timesnet} learns the temporal variations in the 2D space to enhance the representation capability by capturing variations both within individual periods and across multiple periods in the time series. It analyzes the time series from a new dimension of multi-periodicity by extending the analysis of temporal variations from 1D into 2D space. By transforming the 1D time series into a set of 2D tensors, TimesNet breaks the bottleneck of representation capability in the original 1D space, enabling well-performing vision backbones applicable to the transformed time series. Recently, \citet{P105_zhong2023multi} propose a novel multi-scale temporal patching approach to divide the input time series into non-overlapping patches along the temporal dimension in each layer. It treats the time series as patches. FITS \cite{P106_xu2023fits} conducts interpolation in the frequency domain to extend time series and incorporates a low-pass filter to ensure a compact representation while preserving essential information. 

To address tasks involving irregular time series, some input space transformation techniques have been proposed. SplineNet\,\cite{P27_bilovs2022irregularly} generates splines from the input time series and directly utilizes them as input for a neural network. This approach introduces a learnable spline kernel to effectively process the input spline. MIAM\,\cite{P44_lee22multiview} considers multiple views of the input data, including time intervals, missing data indicators, and observation values. These transformed input data are then processed by the multi-view integration attention module to solve downstream tasks. 

\begin{summaryBox}
The approaches for improving data quality collectively highlight a significant trend: \emph{moving beyond the raw time series to create more informative inputs}. Whether through intelligent sample selection, principled decomposition into constituent parts, or transformation into other modalities like images, these methods aim to either distill the most salient information or re-frame the problem to leverage powerful, pre-existing tools. This signifies a maturation in the field, where engineering the data is seen as a critical lever for improving representation quality.
\end{summaryBox}

\subsection{Increasing Data Quantity} \label{section:data_quantity}

\subsubsection{Data Augmentation} \label{section:augmentation} Unlike other data types, augmenting time series requires special care because arbitrary perturbations can alter temporal dependencies, multi-scale patterns, inter-variable relationships, or even the semantic label of the signal. Existing methods can be grouped into two broad design patterns: random augmentation, which creates diverse views with simple stochastic transformations, and policy-based augmentation, which constrains or learns the augmentation rule so that the generated view better preserves time-series semantics.

\paragraph{Random Augmentation} A first family applies generic, stochastic perturbations to time-series segments. TS2Vec\,\cite{P88_yue2022ts2vec} randomly treats two overlapping time segments as positive pairs by combining timestamp masking and random cropping, producing a robust contextual representation without injecting transformation- or cropping-invariance biases. Building on this, TimesURL\,\cite{P111_liu2023timesurl} uses frequency-temporal augmentation that combines frequency mixing with random cropping to preserve the temporal property, and TS-CoT\,\cite{P103_zhang2023co} creates diverse views designed to be robust to noisy time series. A second sub-family explicitly targets the frequency domain: TF-C\,\cite{P57_zhang2022self} introduces frequency-domain augmentation by randomly adding or removing frequency components, exposing the model to a range of spectral variations.


\paragraph{Policy-based Augmentation} Instead of relying on randomness, several studies devise specific criteria for augmentation; we group them by the underlying mechanism. \textbf{(i) DTW-based augmentation}. TimeCLR\,\cite{P81_yang2022timeclr} induces phase shifts and amplitude changes via dynamic time warping while preserving structural information, and TempCLR\,\cite{P71_yang2023tempclr} uses a DTW-based negative-sampling strategy at the sequence level for video-paragraph alignment. \textbf{(ii) Frequency-domain augmentation}. \citet{P99_demirel2023finding} propose a mixup variant for non-stationary quasi-periodic time series that blends magnitude and phase of frequency components to avoid destructive interference; MF-CLR\,\cite{P129_duan2024mfclr} uses a Dual-Twister scheme that injects noise along both dimensions of multi-frequency time series; and \citet{P132_li2024unicl} employ a unified, trainable augmentation operation that exploits spectral information to reduce bias and improve cross-domain generalization. \textbf{(iii) Context- and structure-preserving augmentation}. BTSF\,\cite{P92_yang2022unsupervised} uses standard dropout as an instance-level augmentation to preserve global temporal information, \citet{P98_shin23e} propose context-attached augmentation that appends preceding and succeeding instances rather than perturbing the target itself, and RIM\,\cite{P54_aboussalah2023recursive} generates additional samples through a recursive interpolation that controls the deviation from the original trajectory. \textbf{(iv) Adaptive/learnable augmentation}. InfoTS\,\cite{P76_luo2023time} selects augmentations on the fly using information-aware criteria for high-fidelity, diverse positives; AutoTCL\,\cite{P118_zheng2024parametric} separates each instance into informative and task-irrelevant parts via a learned factorization and only transforms the latter; and UniCL's\,\cite{P132_li2024unicl} spectral-information-based augmentation is jointly optimized with the contrastive objective.

For video data, \citet{P22_chen2022frame} sample two overlapped subsequences and treat overlapped timestamps as positive pairs (with Gaussian-weighted neighbors as additional positives); \citet{P25_Zhang_2022_WACV} use joint temporal and spatial augmentations as a regularizer in a contrastive framework; DynaAugment\,\cite{P20_kim2023exploring} dynamically changes augmentation magnitude over time via Fourier sampling to learn temporal variation; and FreqAug\,\cite{P23_kim2023frequency} stochastically eliminates spatial or temporal low-frequency components to push the model toward more dynamic features.

\textbf{\emph{Takeaway}}. Data augmentation for time series is moving from generic random transformations toward semantics-preserving and adaptive policies. The central design question is no longer only how to create more samples, but how to create alternative views that remain faithful to the original temporal meaning while exposing the model to useful variability.

\subsubsection{Sample Generation and Curation} \label{section:sample_generation} Sample generation is a popular technique that increases the size and diversity of the training data when the data is scarce. It explicitly creates new samples by transformation or generative models. For enhancing the noise resilience, \citet{P34_nguyen2023learning} propose a novel noise-resilient sampling strategy by exploiting a parameter-free discrete wavelet transform low-pass filter to generate a perturbed version of the original time series. By leveraging the LLM, LAVILA\,\cite{P35_zhao2023learning} learns better video-language embedding when only few text annotations are available. Using the available video-text data, it first fine-tunes LLM to generate text narration given visual input. Then, densely annotated videos via a fine-tuned narrator are used for video-text contrastive learning. Regarding curation, \citet{P127_liu2024timer} introduce a large-scale unified time-series dataset, encompassing seven domains with up to 1 billion time points. Similarly, another study \cite{P131_goswami2024moment} present a comprehensive benchmark called Time Series Pile, which aggregates a diverse collection of publicly available datasets across 13 unique domains, comprising 13 million unique time series and 1.23 billion timestamps. These extensive datasets are expected to to play a crucial role in training large-scale time-series models that can be transferred to various data-scarce scenarios.

\begin{summaryBox}
The methods for increasing data quantity underscore two parallel and crucial trends for the future of time-series representation learning. The first is the evolution of data augmentation from simple, random transformations to more sophisticated, learnable strategies that better preserve semantic integrity. The second is the dual effort of using powerful generative models to create high-quality synthetic data while also curating massive, real-world benchmark datasets. Together, these trends are \emph{laying the essential groundwork for training the next generation of large-scale, general-purpose time-series foundation models}.
\end{summaryBox}
\begin{figure}[!t]
    \centering
    \includegraphics[width=\linewidth]{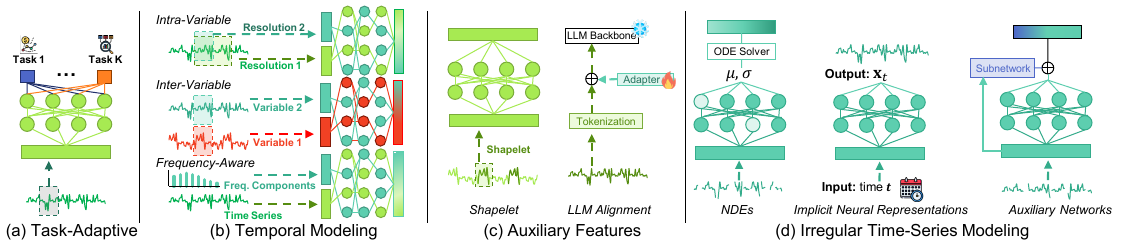}
    \vspace*{-0.7cm}
    \caption{Taxonomic gallery of neural architectural approaches. This figure expands only the network-architecture block of the overall framework in Fig.~\ref{figure:paper_structure} by illustrating task-adaptive submodules, temporal modeling, auxiliary-feature extraction, and continuous or irregular time-series modeling.} \label{figure:neural_architectures}
    \Description{Illustrative examples of neural architectural approaches.}
    \vspace*{-0.5cm}
\end{figure}

\section{Neural Architectural Approaches} \label{section:architectures}
As neural architectures play a crucial role in the quality of representations\,\cite{trirat2020experimental}, this section examines novel \emph{network-architecture designs} aimed at enhancing representation learning. Within the overall framework of Fig.~\ref{figure:paper_structure}, the network-architecture element is the encoder $f_e$ that maps the (possibly augmented) training data (\S\ref{section:data_centric}) into the representation space, where it is shaped by the learning objectives (\S\ref{section:learning_method}). The sub-categories that populate this element are illustrated in Fig.~\ref{figure:neural_architectures}.

\subsection{Task-Adaptive Submodules} \label{section:task_module} 

\citet{P01_chen2021deep} propose multi-task representation learning method called MTRL by exploiting supervised learning for classification and unsupervised learning for retrieval. MTRL jointly optimizes the two downstream tasks via a combination of deep wavelet decomposition networks to extract multi-scale subseries and 1D-CNN residual networks to learn time-domain features, thus improving the performance of downstream tasks. Another method using the wavelet transform, called WHEN\,\cite{P96_WHEN_KDD23}, newly designs two types of attention modules: WaveAtt and DTWAtt. In the WaveAtt module, the study proposes a novel data-dependent wavelet function to analyze dynamic frequency components in non-stationary time series. In the DTWAtt module, WHEN transforms the DTW technique into the form of the DTW attention. Here, all input sequences are synchronized with a universal parameter sequence to overcome the time distortion problem in multiple time series. Then, the outputs from the new modules are further combined with task-dependent neural networks to perform the downstream tasks, such as forecasting. \citet{P89_liang2023units} introduce UniTS, which uses a pre-training module that consists of templates from various self-supervised learning methods. Subsequently, the pre-trained representations are fused. Then, the results of feature fusion are applied to task-specific output models. Due to the proliferation of edge devices, a study\,\cite{P97_SparseTrans_KDD23} proposes a novel model compression technique to make lightweight Transformers for multivariate time-series problems using network pruning, weight binarization, and adaptation of attention modules that can substantially reduce both model size\,(i.e., \# parameters) and computational complexity\,(i.e., FLOPs). The paper demonstrates that compressed Transformers using the proposed technique achieve comparable accuracy to their original counterparts\,\cite{P03_TST_KDD} despite the substantial reduction. These compressed neural networks have the potential to enable DL-based models across new applications and smaller computational environments.

Recently, \citet{P123_gao2024units} present an innovative approach to time-series analysis that unifies various tasks within a single, adaptable model. This model leverages a token-based architecture inspired by LLMs, enabling it to handle diverse time series tasks without requiring task-specific modules. The design's strength lies in its use of sequence tokens, prompt tokens, and task tokens, which provide context and instructions, allowing the model to quickly adapt to new tasks across different domains and datasets without the need for fine-tuning. Similarly, \citet{P125_zhang2024upme} propose a pretraining-based encoder-decoder network with sparse dependency graph construction and temporal-channel layers. A sparse dependency graph is constructed to capture the dependencies between different channels in the multivariate data. The temporal-channel layers sit between the frozen pre-trained encoder and the decoder. These layers are composed of a standard Transformer layer combined with a Graph Transformer layer, which takes the sparse dependency graph as input. This allows the model to capture both temporal and cross-channel dependencies more effectively during fine-tuning for different downstram tasks. MOMENT\,\cite{P131_goswami2024moment} leverages transformer architectures and adapts them for time series tasks by using techniques like masking and patching to manage varying time-series lengths and complexities. MOMENT models are versatile and can be fine-tuned for various tasks by using a different projection head, e.g., reconstruction or forecasting.

\begin{summaryBox}
The studies here signify a broader trend in representation learning: \emph{the shift from monolithic, task-specific models to a more flexible pre-train and fine-tune paradigm}. The core idea is to learn a universal, task-agnostic encoder and then adapt it to specific downstream tasks by attaching small, lightweight modules. This approach offers a better balance between generalization from pre-training and specialization for a given task, promoting modularity and efficiency.
\end{summaryBox}

\subsection{General Temporal Modeling} \label{section:general_temporal_modeling}

\subsubsection{Intra-Variable Modeling} Studies in this group capture patterns and dependencies within each time-series variable. We classify them into long-term modeling and multi-scale or multi-resolution modeling, depending on whether the main goal is to enlarge the temporal receptive field or to represent temporal patterns at several granularities.

\paragraph{Long-Term Modeling} A unifying theme among long-term encoders is enlarging the effective receptive field so that representations can integrate distant time steps. This goal is pursued through three architectural strategies. First, dilated and modernized convolutional designs (e.g., \citet{P91_franceschi2019unsupervised}'s dilated causal 1D-CNN encoder, CoInception\,\cite{P34_nguyen2023learning}'s dilated Inception backbone with skip connections, and ModernTCN\,\cite{P124_donghao2024moderntcn}'s depth-wise design) explicitly enlarge the receptive field while preserving the parallelism of CNNs. Second, Transformer-based designs such as TST\,\cite{P03_TST_KDD} use multi-head self-attention so that each time step can attend to its full past–future context, with multiple heads capturing different representation subspaces. Third, state-space and memory-augmented designs---including SpaceTime\,\cite{P17_zhang2023effectively}, which redesigns encoder–decoders as companion-matrix state-space models, and MemDPC\,\cite{P38_han2020memory}, which augments dense predictive coding with an external memory---produce a global view of the sequence with linear-time updates. \citet{P13_tonekaboni2022decoupling} complement these architectural choices with a counterfactual regularizer that explicitly decouples global and local factors.

\paragraph{Multi-Scale and Multi-Resolution Modeling} Multi-scale methods explicitly represent temporal patterns at different granularities. MSD-Mixer\,\cite{P105_zhong2023multi} uses multi-scale patching and MLP mixing to capture intra- and inter-patch variations, while \citet{P108_fraikin2023t} attach time-embedding modules to learn trend, periodicity, and distribution-shift information. CoInception\,\cite{P34_nguyen2023learning}, COMET\,\cite{P100_wang2023contrast}, CARD\,\cite{P121_wang2024card}, and TSLANet\,\cite{P126_eldele2024tslanet} also capture multiple temporal scales through multi-scale filters, hierarchical medical-time-series blocks, token blending, or adaptive spectral blocks. In video and satellite-image time series, T-C3D\,\cite{P02_liu2020real}, \citet{P72_sener2020temporal}, TCGL\,\cite{P68_liu2022tcgl}, and \citet{P30_sanchez2019learning} extend this idea to high-dimensional temporal data by combining temporal encoding, non-local aggregation, graph contrastive learning, or VAE-GAN-based disentanglement.

\textbf{\emph{Takeaway}}. Intra-variable modeling has evolved from strictly sequential modeling toward architectures with broader and more flexible receptive fields. The dominant pattern is to combine global temporal context with multi-resolution structure so that the learned representation can encode both long-range dependencies and local temporal variations.

\subsubsection{Inter-Variable Modeling} Approaches in this group are designed to explicitly capture relationships and dependencies \emph{between} variables or channels in a multivariate time series. While these methods inherently perform intra-variable modeling, their primary novelty lies in how they model the interactions between variables. \citet{P66_guo2021ssan} introduce a separable self-attention\,(SSA) module designed to capture spatial and temporal correlations in videos separately. The proposed SSA improves the understanding of actions in videos and demonstrating superior performance in action recognition and video retrieval tasks. For time series domains that exhibit a common causal structure but possess varying time lags, SASA\,\cite{P77_cai2021time} aligns the source and target representation spaces by establishing alignment on intra-variable and inter-variable sparse associative graphs. MARINA\,\cite{P37_xie2022marina} comprises a temporal module learning temporal correlations using MLP and residual connections alongside a spatial module capturing spatial correlations between time-series data using GAT. It comprehends relationships among various variables and grasps intricate patterns within time series data, thereby achieving universal and flexible time-series representation learning. \citet{P122_wang2023fully} propose a fully-connected spatial-temporal graph neural network\,(FC-STGNN) to model the spatio-temporal dependencies of multivariate time-series data. FC-STGNN consists of a fully-connected graph to model correlations between various sensors and a moving-pooling GNN layer to capture local temporal patterns. MSD-Mixer\,\cite{P105_zhong2023multi} introduces a novel temporal patching technique that breaks down the time series into multi-scale patches, helping the model to better capture both intra- and inter-patch variations and correlations between different channels. 

\citet{P119_xiao2023gaformer} propose a novel positional embedding, group embedding, which assigns input instances to a set of learnable group tokens to embed instance-specific inter-channel relationships and temporal structures. Grouping occurs in two sequential transformers from channel-wise and temporal perspectives. HierCorrPool\,\cite{P46_wang2023multivariate} is a new framework that captures both hierarchical correlations and dynamic properties by using a novel hierarchical correlation pooling scheme and sequential graphs. A recent work, CARD\,\cite{P121_wang2024card}, is proposed to effectively capture dependencies across multiple channels (variables) by incorporating channel alignment that allows the model to share information among different channels, improving the ability to capture inter-dependencies. \citet{P124_donghao2024moderntcn} propose a modernized TCN model by separating the processing of temporal and feature information, which is a departure from traditional CNNs that typically mix these aspects together. This separation is achieved through depth-wise convolution and convolutional feed-forward networks. \citet{P125_zhang2024upme} introduce a new pre-training framework, UP2ME, that constructs a dependency graph among channels to capture cross-channel relationships when the pre-trained model is fine-tuned on multivariate time series. UP2ME incorporates learnable temporal-channel layers that adjust both temporal and cross-channel dependencies.

\subsubsection{Frequency-Aware Aggregation} Frequency-aware aggregation treats spectral structure as an explicit source of temporal information. A recent survey \cite{DL_Freq_TS_Survey_KDD_25} provides a focused review of frequency-transformation-based DL for time series. In contrast, our discussion positions frequency-aware modeling as one architectural strategy within the broader taxonomy of universal time-series representation learning. \citet{P45_wang2018multilevel} propose a wavelet-based neural architecture, called mWDN, by integrating multi-level discrete wavelet decomposition into existing neural networks for building frequency-aware deep models. This integration enables the fine-tuning of all parameters within the framework while preserving the benefits of multi-layer discrete wavelet decomposition in frequency learning. \citet{P92_yang2022unsupervised} propose an unsupervised representation learning framework for time series, named BTSF. BTFS enhances the representation quality through the more reasonable construction of contrastive pairs and the adequate integration of temporal and spectral information. BTSF constitutes an iterative application of a novel bi-linear temporal-spectral fusion, explicitly encoding affinities between time and frequency pairs. To adequately use the informative affinities, BTSF further uses a cross-domain interaction with spectrum-to-time and time-to-spectrum aggregation modules to iteratively refine temporal and spectral features for cycle update, proven effective by empirical and theoretical analysis.

Another recent method\,\cite{P96_WHEN_KDD23} introduces a data-dependent wavelet function within a BiLSTM network to analyze dynamic frequency components of non-stationary time series. \citet{P113_zhou2023one} design a frequency adapter using fast Fourier transform to capture the frequency domain based on a pre-trained language model. TimesNet\,\cite{P82_wu2023timesnet} introduces TimesBlocks, using Fourier transform to extract periods and Inception blocks for efficient parameter extraction, followed by adaptive aggregation using amplitude values. \citet{P106_xu2023fits} propose FITS, a lightweight yet powerful model for time-series analysis, which extends time series segments by interpolating in the complex frequency domain. \citet{P126_eldele2024tslanet} integrate an adaptive spectral block, leveraging Fourier analysis to enhance feature representation and effectively handle noise through adaptive thresholding. The authors also employ an interactive convolution block to improve its ability to decode complex temporal patterns. MF-CLR\,\cite{P129_duan2024mfclr} introduces a novel approach, especially focusing on datasets where the data is collected at multiple frequencies, such as in financial markets. MF-CLR has a hierarchical mechanism that processes different frequency components of time-series data separately. It creates embeddings by contrasting subseries with adjacent frequencies, ensuring that the model can capture the relationships between different frequency bands effectively.

\begin{summaryBox}
The evolution in general temporal modeling is characterized by a dual pursuit: \emph{capturing dependencies over longer time spans and across multiple scales}. Architecturally, this marks a clear progression from sequential models\,(e.g., RNNs), which struggle with long-term memory, towards architectures with a global receptive field, such as Transformers, and efficient long-convolution models\,(e.g., TCNs). Concurrently, there is a growing recognition that integrating frequency-domain analysis directly into the network provides a more holistic view of temporal dynamics, leading to more robust and versatile representations.
\end{summaryBox}

\subsection{Auxiliary Feature Extraction} \label{section:aux_feature}
Instead of refining the core temporal or inter-variable architecture as discussed above, methods in this category primarily focus on enriching the time-series representation by incorporating extra information or an additional semantic (i.e., contextual) representation. This includes techniques such as transforming time series into shapelets or images to leverage vision models, or aligning time-series representations with the semantic space of an LLM. \citet{LLM_TS_Survey_IJCAI_24} provide a dedicated survey of LLMs for time series. Here, we discuss this direction more narrowly as an auxiliary-feature strategy by aligning numerical signals with semantic spaces learned from language, vision, or multi-modal foundation models.

\subsubsection{Shapelet and Motif Modeling} \citet{P07_liang2023contrastive} propose CSL, a unified shapelet-based encoder with multi-scale alignment, to transform raw multivariate time series into a set of shapelets and learn the representations using the shapelets. \citet{P115_qu2024cnn} demonstrate that shapelets, traditionally used for time-series modeling, are equivalent to specific CNN kernels that involves a squared norm and pooling operation. The authors propose a novel CNN layer called ShapeConv, where the kernels act as shapelets, enabling high interpretability with shaping regularization.

\subsubsection{Contextual Modeling and LLM Alignment} Audio word2vec\,\cite{P05_chen2019audio} extends vector representations to consider the sequential phonetic structures of the audio segments trained with speaker-content disentanglement based segmental sequence-to-sequence autoencoder. For graph time series, STANE\,\cite{P83_liu2019towards} guides the context sampling process to focus on the crucial part of the data in the graph attention networks. \citet{P86_rahman2021tribert} introduce a tri-modal VilBERT-inspired model by integrating separate encoders for vision, pose, and audio modalities into a single network. DelTa\,\cite{P14_anand2021delta} uses 2D images of time series such that pre-trained models on large image datasets can be used. DelTa proposes two versions of using pre-trianed vision models: layout aligned version and layout independent version. \citet{P95_lee2022weakly} present a novel BMA-Memory framework for bimodal representation learning, focusing on sound and image data. This memory allows for the association of features between different modalities, even when data pairs are weakly paired or unpaired.  

Following TST\,\cite{P03_TST_KDD}, \citet{P67_chowdhury2022tarnet} propose TARNet to reconstruct important timestamps using a newly designed masking layer to improve downstream task performance. It decouples data reconstruction from the downstream task and uses a data-driven masking strategy instead of random masking via self-attention score distribution generated by the transformer encoder during the downstream task training to determine a set of important timestamps to mask. \citet{P21_kim2023feat} introduce another MLP-based feature-wise encoder together with a element-wise gating layer built on top of TS2Vec\,\cite{P88_yue2022ts2vec}, i.e., feature-agnostic temporal representation using TCN, to flexibly learn the influence of feature-specific patterns per timestamp in a data-driven manner. \citet{P42_choi2023multi} also extend the TS2Vec encoder\,\cite{P88_yue2022ts2vec} with a multi-task self-supervised learning framework by combining contextual, temporal, and transformation consistencies into a single networks. 

One Fits All\,\cite{P49_zhou2023one} uses a pre-trained language model\,(e.g., GPT-2) by freezing self-attention and feed-forward layers and fine-tuning the remaining layers. Input embedding and normalization layers are modified for time-series data. By doing so, it can benefit from the universality of the Transformer models on time-series data. Based on a pre-trained language model, \citet{P113_zhou2023one} additionally design task-specific gates and adapters. These adapters allow the model to effectively leverage the generalization capabilities of the pre-trained LM while adapting to the specific demands of various time series tasks. For example, temporal adapters focus on modeling time-based correlations, channel adapters on handling multi-dimensional data, and frequency adapters on capturing global patterns through Fourier transforms. DTS\,\cite{P84_li2022towards} is a disentangled representation learning framework that operates through two interoperating components: (1) an individual factor disentanglement module that decomposes latent representations into independent semantic factors, and (2) a group segment disentanglement module that groups these factors to learn higher-level, compositional semantic patterns, creating a hierarchical and interpretable representation.

NuTime\,\cite{P107_lin2023nutime} introduces a novel approach to pre-training models on large-scale time series by leveraging a numerically multi-scaled embedding\,(NME) technique. The model starts by partitioning the data into non-overlapping windows, each represented by three components: its normalized shape, mean, and standard deviation. These components are then concatenated and transformed into tokens suitable for a Transformer encoder. By considering all possible numerical scales, NME enables the model to effectively handle scalar values of arbitrary magnitudes within the data. This technique ensures the smooth flow of gradients during training, making the model highly effective for large-scale time series data. UniTTab\,\cite{P50_luetto2023one} is a Transformer-based framework for time-dependent heterogeneous tabular data. It uses row-type dependent embedding and different feature representation methods for categorical and numerical data, respectively. \citet{P132_li2024unicl} employ an LLM-based encoder initialized with pre-trained weights from the text encoder of CLIP\,\cite{CLIP} designed to maintain variable independence to address the challenge of embedding time-series data from multiple domains without introducing domain-specific biases. This approach enables the pre-trained models to be highly adaptable and effective across various time series tasks.

\begin{summaryBox}
The key trend in this area is the leveraging of powerful, pre-trained foundation models from other domains---primarily vision and language---to enrich time-series representations. \emph{This has evolved from transforming time series into images to use vision models to, more recently, aligning time-series representations with the rich semantic space of LLMs}. This alignment enables powerful capabilities and brings a new level of contextual understanding to the time-series domain.
\end{summaryBox}

\subsection{Continuous Temporal and Irregular Time-Series Modeling} \label{section:cons_temporal}

\subsubsection{Neural Differential Equations (NDE)} ODE-RNN\,\cite{P28_RubanovaCD19} is an early attempt applying neural ordinary differential equations\,(Neural ODE) for time series. It serves as an encoder in a latent ODE model, facilitating interpolation, extrapolation, and classification tasks for ISTS. ANCDE\,\cite{P04_jhin2021attentive} and EXIT\,\cite{P19_jhin2022exit} propose neural controlled differential equation (Neural CDE)-based approaches for classification and forecasting with ISTS. ANCDE leverages two Neural CDEs, one for learning attention from the input time series and another for creating representations for downstream tasks. Likewise, EXIT uses Neural CDEs as part of the encoder-decoder network, enabling interpolation and extrapolation of ISTS. Also, CrossPyramid\,\cite{P11_abushaqra2022crosspyramid} addresses the limitation of ODE-RNN which is the high dependence on the initial observations by using pyramid attention and cross-level ODE-RNN. Contiformer\,\cite{P101_chen2023ContiFormer} merges NDEs into the attention mechanism of a transformer by modeling key and value with Neural ODEs. \citet{P114_oh2024stable} enhance the stability and performance of neural stochastic differential equations (Neural SDEs) when applied to ISTS with three distinct classes of Neural SDEs. Each class features unique drift and diffusion functions designed to address the challenges of stability and robustness in the stochastic modeling of ISTS. The authors also emphasize the importance of well-defined diffusion functions to prevent issues like stochastic destabilization.

\subsubsection{Implicit Neural Representations (INR)} HyperTime\,\cite{P26_fons2022hypertime} introduces INR for time series used for imputation and reconstruction by taking timestamps as input and outputting the original time series. It consists of two networks Set Encoder and HyperNet Decoder. \citet{P75_naour2023time} introduce TimeFlow to deal with modeling issues such as irregular time steps. With a hyper-network, the framework modulates INR that is a parameterized continuous function on multiple time series.

\subsubsection{Auxiliary (Sub-)Networks} LIME-RNN\,\cite{P18_ma2019end} introduces a weighted linear memory vector into RNNs for time-series imputation and prediction. Another work by \citet{P33_bianchi2019learning} proposes a temporal kernelized autoencoder to learn representations aligned to a kernel function which are designed for handling missing values. mTAN\,\cite{P43_shukla2021multitime} learns the representation of continuous time values by applying the attention mechanism to ISTS. The key innovation of mTANs lies in its continuous-time attention mechanism, which generalizes positional encoding typically used in transformers to operate in continuous time rather than discrete steps. This mechanism leverages multiple time embeddings to flexibly capture both periodic and non-periodic patterns in the data, allowing the network to produce fixed-length representations of time series regardless of the number or irregularity of observations. TE-ESN\,\cite{P70_SunHSCSC021} employs a requisite time encoding mechanism to acquire knowledge from ISTS, with the representation being learned within echo state networks. 

Continuous recurrent units\,\cite{P39_schirmer2022modeling} update hidden states based on linear stochastic differential equations, which are solved by the continuous-discrete Kalman filter. Based on continuous-discrete filtering theory, \citet{P47_ansari23a} introduce a neural continuous-discrete state space model that employs a set of auxiliary variables as an intermediate representation. This sub-network of auxiliary variables disentangles the recognition of observations from the system's core dynamics, enabling principled Bayesian inference for the primary dynamic states. \citet{P40_bilos23a} suggest a representation learning approach based on denoising diffusion models adapted for ISTS with complex dynamics. By gradually adding noise to the entire function, the model can effectively capture underlying continuous processes. TriD-MAE\,\cite{P102_zhang2023trid} is a pre-trained model based on TCN blocks with attention scale fusion to handle ISTS. \citet{P133_senane2024self} propose a novel architecture called TSDE specifically designed to handle ISTS with dual-orthogonal Transformer encoders that are integrated with a crossover mechanism. This structure processes the observed segments of the time series, which are divided by an imputation-interpolation-forecasting mask. The architecture then conditions a reverse diffusion process on these embeddings to predict and correct noise in the masked parts of the series, allowing TSDE to generate robust representations, making it particularly effective for irregular and noisy time series.

\begin{summaryBox}
The overarching trend for irregular time series is \emph{the shift from discrete-time assumptions to continuous-time modeling}. Rather than forcing data onto a fixed grid via imputation or sampling, methods based on NDE and INR learn a continuous function of time. This provides a more principled and natural way to handle the challenges of irregular sampling and missing data, representing a fundamental shift in how we model temporal processes.
\end{summaryBox}

\begin{figure}[!t]
    \centering
    \includegraphics[width=\linewidth]{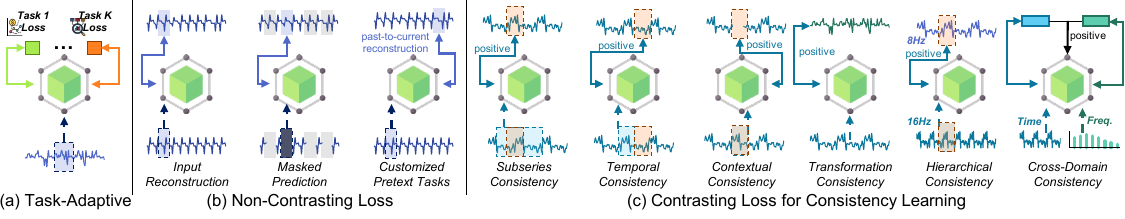}
    \vspace*{-0.7cm}
    \caption{Taxonomic gallery of learning-focused approaches. This figure expands only the learning-objective block of the overall framework in Fig.~\ref{figure:paper_structure} by illustrating task-adaptive objectives, non-contrasting temporal-dependency objectives, and contrastive consistency objectives.} \label{figure:learning_focused}
    \Description{Illustrative examples of learning-focused approaches.}
    \vspace*{-0.5cm}
\end{figure}

\vspace*{-0.1cm}
\section{Learning-Focused Approaches} \label{section:learning_method}
Studies in this category center on devising \emph{novel learning objectives} for the representation-learning process, i.e., model (pre-)training. Within the overall framework of Fig.~\ref{figure:paper_structure}, the learning-objective element is what shapes the encoder (\S\ref{section:architectures}) once it has consumed the training data (\S\ref{section:data_centric}), and ultimately determines what kind of structure the representation space acquires. Fig.~\ref{figure:learning_focused} unpacks this element into three groups: task-adaptive, non-contrasting, and contrasting losses.

\subsection{Task-Adaptive Objective Functions} \label{section:task_loss}

\citet{P18_ma2019end} develop a set of task-specific loss functions to train the LIME-RNN with incomplete time series in an end-to-end way, simultaneously achieving imputation and prediction. TriBERT\,\cite{P86_rahman2021tribert} designs separate losses for a specific modality. Weakly supervised classification for vision-based modalities and classification loss for audio modality. \citet{P55_hadji2021representation} use a soft version of DTW\,(soft-DTW) to compute the loss between two videos with the same class in the weakly-supervised setting. Cycle-consistency is forced by matching two DTW results ($X \to Y$ and $Y \to X$) as the same. \citet{P105_zhong2023multi} propose MSD-Mixer with a novel loss function to constrain both the magnitude and auto-correlation of the decomposition residual used together with the supervised loss of the target downstream task during training. This loss function facilitates MSD-Mixer to extract more temporal patterns into the components to be used for the downstream task. \citet{P121_wang2024card} propose a robust loss function designed to mitigate over-fitting by weighting forecasting errors based on prediction uncertainties, which contributes to more accurate and robust time-series forecasting and anomaly detection. UniTS\,\cite{P123_gao2024units} unifies multiple tasks within a single model framework by using a universal task specification that allows a single model to handle diverse downstream tasks without the need for task-specific modules. It employs a prompting-based framework to convert different tasks into a unified token representation. This unified approach with a single set of shared weights can generalize better and perform multiple tasks simultaneously.

\begin{summaryBox}
The trend in this category is the move towards greater flexibility and generalization through the design of loss functions. Instead of relying on a single, fixed objective, methods like UniTS use prompting frameworks to specify different downstream tasks within a unified loss structure. This allows a single model to be trained for multiple purposes, \emph{representing a shift from specialized models to more general and adaptable learning paradigms}.
\end{summaryBox}

\vspace*{-0.15cm}
\subsection{Non-Contrasting Losses for Temporal Dependency Learning} \label{section:non_cl}

\subsubsection{Input Reconstruction} Sqn2Vec\,\cite{P65_nguyen2019sqn2vec} learns low-dimensional continuous feature vectors for sequence data by predicting singleton symbols and sequential patterns to satisfy a gap constraint. \citet{P05_chen2019audio} introduce unsupervised training of audio word2vec using a sequence-to-sequence autoencoder on unannotated audio time series. The encoder includes a segmentation gate trained with reinforcement learning to represent utterances as vectors carrying phonetic information. Wave2Vec\,\cite{P94_yuan2019wave2vec} jointly models inherent and temporal representations of biosignals and provides clinically meaningful interpretation with enhanced interpretability. \citet{P30_sanchez2019learning} propose a method for satellite image time series by combining VAE and GAN to learn image-to-image translation, where one image in a time series is translated into another. 

Ti-MAE\,\cite{P74_li2023_TiMAE}, a masked autoencoder framework, addresses the distribution shift problem by learning strong representations with less inductive bias or hierarchical trick. Its masking strategy creates different views for the encoder in each iteration, fully leveraging the whole input time series during training. \citet{P112_chen2024multi} use probabilistic masked autoencoding where segment-wise masking schemes and rate-aware positional encodings are devised to enable the characterization of multi-scale temporal dynamics. In the pre-training phase, the encoders generate rich and holistic representations of multi-rate time series. A temporal alignment mechanism is devised to refine synthesized features for dynamic predictive modeling through feature block division and block-wise convolution. \citet{P117_lee2024learning} argue that embedding time series patches independently rather than capturing dependencies between them can lead to better representation learning and introduce a simple patch reconstruction task, where each time series patch is autoencoded without considering other patches. This task helps in learning meaningful representations for each patch independently.

\subsubsection{Masked Prediction} TST\,\cite{P03_TST_KDD} adopts an encoder-only Transformer network with a masked prediction (denoising) objective. The losses of TST are computed only from the masked parts or timestamps. Specifically, TST trains the Transformer encoder to extract dense vector representations of multivariate time series using the denoising objective on randomly masked time series. Similar to TST, TARNet\,\cite{P67_chowdhury2022tarnet} improves downstream task performance by learning to reconstruct important timestamps using a data-driven masking strategy. The masked timestamps are determined by self-attention scores during the downstream task training. This reconstruction process is trained alternately with the downstream task at every epoch by sharing parameters in a single network. It also enables the model to learn the representations via task-specific reconstruction, which results in improved downstream task performance. \citet{P40_bilos23a} deal with ISTS by treating time-series data as discretization of continuous functions and using denoising diffusion models. By masking some parts of the continuous function and predicting the rest, it captures the temporal dependencies of time-series data and enables generalization across various types of time series. 

SimMTM\,\cite{P63_dong2023simmtm} reconstructs the original series from multiple masked series with series-wise similarity learning and point-wise aggregation to reveal the local structure of the manifold implicitly. It introduces a neighborhood aggregation design for reconstruction by aggregating the point-wise representations of time series based on the similarities learned in the series-wise representation space. Thus, the masked time points are recovered by weighted aggregation of multiple neighbors outside the manifold, allowing SimMTM to assemble complementary temporal variations from multiple masked time series and improve the quality of the reconstruction. In addition, a constraint loss is proposed to guide the series-wise representation learning based on the neighborhood assumption of the time series manifold. UP2ME\,\cite{P125_zhang2024upme} uses task-agnostic univariate pre-training that involves generating univariate instances by varying window lengths and decoupling channels. These instances are then used for pre-training a masked autoencoder, focusing only on temporal dependencies without considering cross-channel dependencies, which allows it to perform tasks immediately after pre-training by simply formulating them as specific mask-reconstruction problems. \citet{P133_senane2024self} propose a time series diffusion embedding model. It combines a diffusion process with a novel imputation-interpolation-forecasting mask to enhance the learning of time-series representations by segmenting time series into observed and masked parts. 

\subsubsection{Customized Pretext Tasks} Customized pretext tasks design self-supervised prediction problems beyond direct reconstruction or masking. We group recent methods by the conceptual type of pretext task they introduce.

\vspace*{-0.15cm}
\paragraph{Sequential-Pattern and Warping-Based Tasks.} \citet{P53_wu2018random} propose random warping series, a positive-definite kernel built from DTW between original and randomly warped series, and Sqn2Vec\,\cite{P65_nguyen2019sqn2vec} learns sequence vectors by predicting both singleton symbols and gap-constrained sequential patterns. \citet{P95_lee2022weakly} learn the association between sound and image representations under both weakly paired and unpaired regimes. \citet{P104_fang2023learning} contrast prior and dynamic graphs to separate common and instance-specific spatial relations, while T-Rep\,\cite{P108_fraikin2023t} uses time-embedding-based pretext tasks to impose a coherent temporal structure on the latent space. As a representative method in this group, random warping series illustrates how DTW-based supervision yields embeddings that respect the warping geometry of time series; the remaining methods replace the warping operator with sequential-pattern, cross-modal, or time-embedding constructs while keeping the overall idea intact.

\vspace*{-0.15cm}
\paragraph{Next-Token or Next-Patch Prediction.} A growing family adapts the next-token paradigm of LLMs to time series. The representative example is Timer\,\cite{P127_liu2024timer}, which adopts a GPT-style decoder-only architecture pre-trained with next-token prediction over a unified, large-scale time-series corpus. TEST\,\cite{P109_sun2023test} tokenizes time series into LLM-readable text-form embeddings, \citet{P128_bian2024multipatch} train LLM backbones with multi-patch prediction in a two-stage curriculum, and TimeSiam\,\cite{P130_dong2024timesiam} reconstructs the masked portion of the current subseries from past observations and adds lineage embeddings to discriminate temporal distances. The shared design principle is that predicting the next token or patch from a long context forces the encoder to model long-range temporal dynamics rather than mere local reconstruction.

\vspace*{-0.15cm}
\paragraph{Video-Specific and Temporal-Alignment Tasks.} A third group treats two views of the same activity as the supervisory signal. \citet{P60_wang2020self} classify the pace of a clip into five categories (super-slow to super-fast). \citet{P29_haresh2021learning} use soft-DTW for cross-video alignment. \citet{P61_wang2021self} use motion-aware curriculum learning over rough spatial partitions. \citet{P58_liang2022self} split clips into three consecutive parts to define continuity-prediction, discontinuity-localization, and missing-section-approximation tasks. Other several methods exploit transformations whose effect on a video is recoverable. CACL\,\cite{P09_guo2022cross} predicts an Edit-distance proxy between a clip and its temporal shuffle; TransRank\,\cite{P85_duan2022transrank} ranks the relative magnitude of transformations applied to a clip; and CSTP\,\cite{P08_zhang2022contrastive} predicts a spatio-temporal overlap rate that bridges a pretext task with downstream contrastive learning. The common design principle is forcing the model to learn ordering, speed, continuity, or transformation intensity. These objectives are especially useful when temporal semantics are expressed through motion rather than through scalar sensor values.

\textbf{\emph{Takeaway}}. Customized pretext tasks show a progression from simple sequence-structure prediction to abstract temporal reasoning. The strongest recent trend is to borrow next-token or next-patch prediction from language modeling while adapting it to the continuity, scale, and semantics of time-series data.

\begin{summaryBox}
The evolution of non-contrasting losses shows a clear trend \emph{towards making the self-supervised task more challenging and abstract to force the model to learn richer features}. This has progressed from simple input reconstruction via autoencoders to more difficult masked prediction tasks, as seen in TST. The most recent development is the adoption of highly abstract, customized pretext tasks, such as next-token prediction in the style of LLMs, which pushes the model to learn complex temporal dynamics beyond mere reconstruction.
\end{summaryBox}

\subsection{Contrasting Losses for Consistency Learning} \label{section:cl}

\subsubsection{Subseries Consistency} \citet{P91_franceschi2019unsupervised} employ the triplet loss\,(i.e., T-Loss) inspired by word2vec\,\cite{word2vec} for learning scalable representations of multivariate time series. It considers a sub-segment belonging to the input time segment as a positive sample to explore sub-series consistency. \citet{P87_TSRep} propose TS-Rep by combining the T-Loss \cite{P91_franceschi2019unsupervised} with the use of nearest neighbors to diversify positive samples. The method is particularly effective in handling the complexity and variability inherent in robot sensor data. \citet{P36_behrmann2021long} separate the representation space into stationary and non-stationary characteristics through contrastive learning from long and short views, enhancing both the video-level and temporally fine-grained tasks. \citet{P48_qian2022temporal} propose a windowing based learning by sampling a long clip from a video and a short clip that lies inside the duration of the long clip. These long and short clips become a positive pair for contrastive learning, and other long clips become negative instances. When conducting contrastive learning, final vectors are made in two different embedding spaces. The first one is a fine-grained space and each embedding is made at each timestamp. The second embedding space is a persistent embedding space and each timestamp embedding is global average pooled for contrastive learning. \citet{P60_wang2020self} select positive pairs from a pair of clips in the same video and negative pair from a pair of clips in different videos. CVRL\,\cite{P64_qian2021spatiotemporal} uses the temporally consistent spatial augmentation and clip selection strategy, where each frames are spatially augmented. Here, two clips in the same video are positive, while two clips in the different videos are negative.

\subsubsection{Temporal Consistency} TNC\,\cite{P90_tonekaboni2021_TNC} leverages temporal neighborhoods as positive samples through the ADF test. By incorporating sample weight adjustment into contrastive loss, the sampling bias problem is alleviated. \citet{P79_ijcai2021_324} incorporate a novel temporal contrasting module to learn temporal dependencies by designing a hard cross-view prediction task that uses past latent features of one augmentation to predict the future of another augmentation for a certain time step. This operation forces the model to learn robust representation by a harder prediction task against any perturbations introduced by different time steps and augmentations. \citet{P59_hajimoradlou2022selfsupervised} introduce similarity distillation along the temporal and instance dimensions for pre-training universal representations. \citet{P81_yang2022timeclr} propose TimeCLR, which enables a feature extractor to learn invariant representations by minimizing the similarity between two augmented views of the same sample. \citet{P122_wang2023fully} integrate the FC graph construction with the moving-pooling GNN. The model is capable of learning high-level features that represent both the spatial and temporal aspects of multivariate time series. This dual focus ensures that the temporal consistency is preserved while also accounting for spatial correlations, which is critical for downstream tasks.

\citet{P32_morgado2020learning} use audio-visual spatial alignment as a pretext task with 360° video data. This pretext task involves spatially misaligned audio and video clips, treated as negative examples for contrastive learning. Besides the non-contrastive loss, \citet{P29_haresh2021learning} jointly use a temporal regularization term (i.e., Contrastive-IDM) to encourage two different frames to be mapped to different points in the embedding space. \citet{P22_chen2022frame} propose sequence contrastive loss to sample two subsequences with an overlap for each video. The overlapped timestamps are considered positives, while the clips from other videos are negatives. Two timestamps neighboring each other also become positive pairs with the Gaussian weight proportional to the temporal distance. A recent study\,\cite{P71_yang2023tempclr} proposes TempCLR to explore temporal dynamics in video-paragraph alignment, leveraging a novel negative sampling strategy based on temporal granularity. By focusing on sequence-level comparison using DTW, TempCLR captures temporal dynamics more effectively. \citet{P41_zhang2023modeling} model videos as stochastic processes by enforcing an arbitrary frame to agree with a time-variant Gaussian distribution conditioned on the start and end frames.

\subsubsection{Contextual Consistency} TimeAutoML\,\cite{P80_jiao2020timeautoml} adopts the AutoML framework, enabling automated configuration and hyperparameter optimization. Negative samples are created by introducing random noise within the range defined by the minimum and maximum values of the given instances. CARL\,\cite{P22_chen2022frame} uses a sequence contrastive loss to learn representations by aligning the sequence similarities between augmented video views. This loss function helps maintain temporal coherence and contextual consistency across frames, making the representations robust to variations in video length and content. TS2Vec\,\cite{P88_yue2022ts2vec} uses randomly overlapped segments to capture multi-scale contextual information through temporal and instance-wise contrastive losses. \citet{P103_zhang2023co} introduce TS-CoT, which is a co-training algorithm that enhances the global consistency of representations from different views. REBAR\,\cite{P120_xu2023retrieval} exploits retrieval-based reconstruction to capture the class-discriminative motifs. This method selects positive samples using the REBAR cross-attention reconstruction trained with a contiguous and intermittent masking strategy. \citet{P98_shin23e} propose a consistency regularization framework based on two overlapping windows and leverage a merged soft label from those two windows as a shared target. \citet{P105_zhong2023multi} propose a novel residual loss to ensure that the model's decomposition leaves only white noise as residual, further contributing to the contextual consistency of the analysis by reducing information loss during the decomposition. PrimeNet\,\cite{P51_chowdhury2023primenet} generates augmented samples based on the observation density and employs a reconstruction task to facilitate the learning of irregular patterns.

\subsubsection{Transformation Consistency} TS-TCC\,\cite{P79_ijcai2021_324} transforms a given time series into two different yet correlated views by weak and strong augmentations. Then, it learns discriminative representations by maximizing the similarity among different contexts of the same sample while minimizing similarity among contexts of different samples, leading to high efficiency in few-labeled data and transfer learning scenarios. RSPNet\,\cite{P56_chen2021rspnet} solves the relative speed perception task and appearance-focused video instance discrimination task using the triplet loss and InfoNCE loss. \citet{P78_jenni2021time} propose a time-equivariant model using a pair of clips as the unit of contrastive learning. If two pairs have same temporal transformation within each pair, then the output of each clip is concatenated for each pair and the concatenated outputs become similar. Auxiliary tasks are also used, e.g., classifying the speed difference between two clips: clip A as 2x and clip B as slow speed.

\subsubsection{Hierarchical and Cross-Scale Consistency} \citet{P88_yue2022ts2vec} propose TS2Vec, which is based on novel contextual consistency learning using two contrastive policies over two augmented time segments with different contexts from randomly overlapped subsequences. Unlike other methods, TS2Vec performs contrastive learning in a hierarchical way over the augmented context views, making a robust contextual representation for each timestamp. Its overall representation can be obtained by max pooling over the corresponding timestamps. By using multi-scale contextual information with granularity, this approach can capture multi-scale contextual information with temporal and instance-wise contrastive losses for the given time series and generate fine-grained representations for any granularity. \citet{P34_nguyen2023learning} design a new loss function by combining ideas from hierarchical and triplet losses. CSL\,\cite{P07_liang2023contrastive} proposes to learn time-series representations shapelets with multi-grained contrasting and multi-scale alignment for capturing information in various time ranges. COMET\,\cite{P100_wang2023contrast} incorporates four-level contrastive losses for medical time series. The overall loss has hyper-coefficients about each loss to compromise the multiple hierarchical contrastive losses. By applying contrastive learning across these multiple levels, the framework can create more robust and comprehensive representations of the data. \citet{P129_duan2024mfclr} leverage a hierarchical structure that captures the different frequencies present in the data and embeds subseries of the time series into two groups based on adjacent frequencies. By enforcing consistency between these groups, it create robust representations that are useful across various frequencies.

CCL\,\cite{P12_kong2020cycle} uses the inclusive relation of a video and frames for a contrastive learning strategy. The video and frames of the inclusive relation are learned to be close to each other in the embedding spaces. \citet{P31_qing2022learning} learn representations on untrimmed video to reduce the amount of labor required for manual trimming and use the rich semantics of untrimmed video. Hierarchical contrastive learning teaches clips that are near in time and topic to be similar. \citet{P25_Zhang_2022_WACV} introduce a model trained with decoupled learning objectives into two contrastive sub-tasks, which are hierarchically spatial and temporal contrast. With graph learning, TCGL\,\cite{P68_liu2022tcgl} uses a spatial-temporal knowledge discovering module for motion-enhanced spatial-temporal representations. It introduces intra- and inter-snippet temporal contrastive graphs to explicitly model multi-scale temporal dependencies via a hybrid graph contrastive learning strategy. TCLR\,\cite{P69_dave2022tclr} is a model for video understanding trained with novel local–local and global–local temporal contrastive losses.

\subsubsection{Cross-Domain and Multi-Task Consistency} FEAT\,\cite{P21_kim2023feat} jointly learns feature-based and temporal consistencies by using hierarchical temporal contrasting, feature contrasting, and reconstruction losses. \citet{P42_choi2023multi} introduce uncertainty weighting approach to weigh multiple contrastive losses by considering homoscedastic uncertainty of multiple tasks, including contextual, temporal, and transformation consistencies. FOCAL\,\cite{P110_liu2023focal} enforces the modality consistency to learn the features shared across modalities and the transformation consistency to learn the modality-specific feature. To also accommodate sporadic deviations from locality due to periodic patterns, temporally close sample pairs and distant sample pairs are constrained by a loose ranking loss. \citet{P57_zhang2022self} argue that time-based and frequency-based representations learned from the same time series should be closer to each other in the time-frequency latent space than representations of different time series. Thus, they introduce time-frequency consistency modeling that aims to minimize the distance between time-based and frequency-based embeddings using a novel consistency loss in the same latent space. TimesURL\,\cite{P111_liu2023timesurl} constructs double Universums as a hard negative, and introduces a joint optimization objective with contrastive learning to capture both segment-level and instance-level information. \citet{P116_lee2024soft} propose SoftCLT with soft contrastive losses for more nuanced learning. It uses soft assignments for an instance-wise contrastive loss based on the distance between time series instances, where a soft assignment determines the degree to which different instances should be contrasted and a temporal contrastive loss that focuses on the differences in timestamps to handle temporal correlations. \citet{P132_li2024unicl} reveal a positive correlation between representation bias and spectral distance in time series, resulting in a variation of contrasting losses optimized to reduce the bias from data augmentation. These losses include the spectral characteristics of time-series data, ensuring that the embeddings generated are more robust and generalizable. The bias is quantified by comparing the difference between the embeddings of the original time series and the augmented versions to reduce the discrepancy. MemDPC\,\cite{P38_han2020memory} involves a predictive attention mechanism applied to a group of compressed memories. This training paradigm ensures that any subsequent states can be synthesized consistently through a convex combination of the condensed representations. \citet{P06_zeng2021contrastive} improve generalization by learning spatially-local/temporally-global and spatially-global/temporally-local features from audio-visual modalities to capture global and local information in a video. This method enables the capturing of both slowly changing patch-level information and fast changing frame-level information. DCLR\,\cite{P16_ding2022dual} presents a dual contrastive formulation by decoupling the input RGB video sequence into two complementary modes, static scene and dynamic motion, to avoid static scene bias.

\begin{summaryBox}
The development of contrastive learning for time series reveals a clear trajectory towards \emph{creating more meaningful positive pairs and more challenging learning signals}. This has evolved from simple consistency between random subseries to more sophisticated notions, including consistency across different temporal scales (hierarchical consistency), between different data domains (time-frequency consistency), and even between different but related tasks. This trend reflects a deeper understanding of what constitutes a semantically similar view of a time series, which is the cornerstone of effective contrastive learning.
\end{summaryBox}

\section{Experimental Design} \label{section:evaluation}
This section describes the typically used experimental design for comparing universal representation learning methods for time series. We describe widely-used protocol and introduce publicly available benchmark datasets with evaluation metrics according to the downstream tasks.

Given a set of $N$ time series $\{(\mathbf{X}_i, \mathbf{y}_i)\}_{i=1}^N$ and $J$ pre-trained representation learning models $\{f_{e,j}\}_{j=1}^J$, this section describes how models are evaluated to determine the best one. As discussed in Section~\ref{section:introduction}, representations of time series play a vital role in solving time-series analysis tasks. We expect that the \emph{learned} representations by $f_e$ generalize to unseen downstream tasks. Accordingly, the most common evaluation method is how learned representations help solve downstream tasks.

Additionally, we need a function $g_d$ that maps a representation (feature) space to a label space. For example, $g_d(f_e(\mathbf{X})): \mathbb{R}^{R \times F} \to \mathbb{R}^{|C|}$ for classification or $g_d(f_e(\mathbf{X})): \mathbb{R}^{R \times F} \to \mathbb{R}^{H}$ for forecasting. This is because $f_e$ is designed to extract feature representations, not to solve the downstream task. Commonly, $g_d$ is implemented as a simple function, such as linear regression, support vector machines, or shallow neural networks because it is enough to solve the downstream task if the learned representations already capture meaningful and discriminative features.


\subsection{Evaluation Procedure}
Let $\mathcal{D} = \{(\mathbf{X}_i, \mathbf{y}_i)\}_{i=1}^N$ denote the downstream data. We then compare the encoders $f_e$ by using the task-specific evaluation metrics of the downstream task. The evaluation procedures are as follows.
\begin{enumerate}[noitemsep]
    \item Train $f_e$ and $g_d$ on the downstream dataset $\mathcal{D}$ with the pre-trained encoder $f_{e,j}$ for each $j$.
    \item Compare task-specific evaluation metric values, e.g., accuracy for classification.
\end{enumerate}
In the first step, there are two common protocols to evaluate the encoders: frozen and fine-tuning.

\subsubsection{Frozen Protocol} As we expect $f_e$ to learn meaningful representations for downstream tasks, we do not update the pre-trained $f_e$, i.e., freezing its weights. Training $g_d$ uses less computation budget and converges faster than the training of $f_e$ and $g_d$ from scratch on the downstream dataset. A common choice of $g_d$ is a linear model\,(e.g., linear regression and logistic regression) or non-parametric method\,(e.g., $k$-nearest neighbors)\,\cite{evaluation_rl_ijcai}. This evaluation approach is usually referred as \emph{linear probing}. To further improve performance with additional computing cost, we can also implement $g_d$ as a nonlinear model, such as shallow networks with a nonlinear activation function.

\subsubsection{Fine-Tuning Protocol} In fine-tuning protocol, we train both pre-trained $f_e$ and $g_d$ as a single model to obtain more performance gain of the downstream task or fill the gap between representation learning and downstream tasks. $g_d$ is also known as a task-specific projection head in the representation learning network. This protocol usually uses small learning rate to optimize the model to preserve the original representation quality.

The combination of linear probing and fine-tuning protocol is also possible, especially for out-of-distribution as the encoder's performance on data sampled from out-of-distribution may degrade after fine-tuning\,\cite{finetuning_underperfrom_ICLR_2022}. Even though the fine-tuning protocol requires more computing budget than the frozen protocol, it empirically performs better than the frozen protocol\,\cite{evaluation_rl_ijcai}.

\smallskip
\noindent
\textbf{End-to-End Protocol}. This protocol is a special case for evaluating a representation learning framework, where both $f_e$ and $g_d$ are trained jointly from scratch for each downstream task \emph{without} pre-training in (1). Notable examples include TimesNet\,\cite{P82_wu2023timesnet}, MSD-Mixer\,\cite{P105_zhong2023multi}, and WHEN\,\cite{P96_WHEN_KDD23}.

\subsection{Benchmark Datasets and Metrics for Downstream Tasks}
We summarize widely-used benchmark datasets and evaluation metrics according to the downstream tasks. Some of these datasets are single-purpose datasets for a particular downstream task, and some are general-purpose time-series datasets that we can use for model evaluation across different tasks. Table~\ref{table:benchmark_summary} presents useful information of the reviewed datasets, including dataset names, dimensions, sizes, application domains, and reference sources.

\subsubsection{Forecasting and Imputation} Since the output of forecasting and imputation tasks are numerical sequences, most studies uses the same benchmark datasets for both tasks. Commonly used datasets are from several application domains and services, including electricity\,(e.g., ETT\,\cite{ETT}), transportation\,(e.g., Traffic\,\cite{P82_wu2023timesnet}), meteorology\,(e.g., Weather\,\cite{P82_wu2023timesnet}), finance\,(e.g., Exchange\,\cite{Exchange}) and control systems\,(e.g., MoJoCo\,\cite{P19_jhin2022exit}). To facilitate the evaluation of forecasting models, a publicly accessible archive is also introduced\,\cite{Monash-TSF}. Given the numerical nature of the predicted results, the most commonly used metrics are mean squared error\,(MSE) and mean absolute error\,(MAE).

\subsubsection{Classification and Clustering} As the classification and clustering tasks both aim to identify the real category to which a time-series sample belongs, existing studies usually use the same set of benchmark datasets to evaluate the model performance. Curated benchmarks comprising heterogeneous time series from various application domains, such as UCR\,\cite{ucr-benchmark} and UEA\,\cite{uea-benchmark}, are the most widely used because they can provide a comprehensive evaluation regarding the generalization of the model being evaluated. Many researchers also use human activity\,(e.g., HAR\,\cite{HAR}) and health\,(e.g., Sepsis\,\cite{Sepsis}) related datasets due to their practicality for real-world applications. We recommend referring to Table~\ref{table:benchmark_summary} for the exhaustive list (including audio and video modalities) of datasets for the classification and clustering tasks.

Regarding the evaluation metrics, while we can evaluate the classification task with accuracy, precision, recall, and F1 score, we usually assess the clustering task with Silhouette score, adjusted rand index\,(ARI), and normalized mutual information\,(NMI) to assess the inherent clusterability due to the absence of label instances. For classification, we may also use the area under the precision-recall curve (AUPRC) to handle the class imbalance cases.

\subsubsection{Regression} Compared to the forecasting and classification tasks, time-series regression, particularly with DL, remains relatively underexplored. Only a handful of public benchmark datasets (e.g., heart rate monitoring data\,\cite{P99_demirel2023finding} and air quality\,\cite{P102_zhang2023trid}) exist. The TSER archive, introduced by \citet{tser_survey}, is the most comprehensive benchmark for time-series regression. Like forecasting and imputation tasks, the metrics for regression are MSE and MAE. Additional metrics, such as root mean squared error\,(RMSE) and R-squared\,($R^2$), are also commonly used.

\subsubsection{Segmentation} Likewise, time-series segmentation with DL is also relatively underexplored. There are two standard curated benchmarks: UTSA\,\cite{UTSA} and TSSB\,\cite{ClaSP}. To assess the segmentation performance, F1 and covering scores are typically used. The F1 score emphasizes the importance of detecting the correct timestamps at which the underlying process changes. In contrast, the covering score focuses on splitting a time series into homogeneous segments and reports a measure for the overlaps of predicted versus labeled segments.

\subsubsection{Anomaly Detection} Anomaly detection is one of the most popular research topics in time series. There are several benchmarks publicly available, as listed in Table~\ref{table:benchmark_summary}. However, as argued by recent studies\,\cite{UCR-TSAD, TimeseAD}, most existing benchmarks are deeply flawed and cannot provide a meaningful assessment of the anomaly detection models. Therefore, we recommend using newly proposed datasets, such as ASD\,\cite{ASD}, TimeSeAD\,\cite{TimeseAD}, and TSB-UAD\,\cite{TSB-UAD}.

Concerning the evaluation metrics, point-adjust F1 score\,\cite{donut} is the most widely used metric for time-series anomaly detection. Nevertheless, this metric is also found to have an overestimation problem, which cannot give a reliable performance evaluation. Accordingly, recent studies\,\cite{DCdetector_KDD, TimeCAD_ECML} have started to adopt more robust evaluation metrics, e.g., VUS\,\cite{vus_vldb}, PA\%K\,\cite{rigorous_tsad}, and eTaPR\,\cite{etapr}.

\begin{table}[H]
\centering
\caption{Summary of public datasets widely used for time-series representation learning. $T$ and $V$ indicate the varying time-series length (or \# frames) and number of variables (or video resolution) per sample, respectively.} \label{table:benchmark_summary}
\resizebox{\textwidth}{!}{%
\begin{tabular}{@{}ccccccc@{}}
\toprule
\textbf{Downstream Tasks} &
  \textbf{Dataset Name} &
  \textbf{Size} &
  \textbf{Dimension} &
  \textbf{Domain} &
  \textbf{Modality} &
  \textbf{Reference Source} \\ \midrule
\multirow{14}{*}{Forecasting \& Imputation} &
  ETTh &
  14,307 &
  7 &
  Electric Power &
  time series &
  \cite{ETT} \\
                              & ETTm                & 57,507       & 7                & Electric Power          & time series    & \cite{ETT}              \\
                              & Electricity         & 26,304       & 321              & Electricity Consumption & time series    & \cite{P82_wu2023timesnet}         \\
                              & Traffic             & 17,451       & 862              & Transportation          & time series    & \cite{P82_wu2023timesnet}         \\
                              & PEMS-BAY            & 16,937,179   & 325              & Transportation          & spatiotemporal & \cite{Traffic}          \\
                              & METR-LA             & 6,519,002    & 207              & Transportation          & spatiotemporal & \cite{Traffic}          \\
                              & Weather             & 52,603       & 21               & Climatological Data     & time series    & \cite{P82_wu2023timesnet}         \\
                              & Exchange            & 7,588        & 8                & Daily Exchange Rate     & time series    & \cite{Exchange}         \\
                              & ILI                 & 861          & 7                & Illness                 & time series    & \cite{P82_wu2023timesnet}         \\
                              & Google Stock        & 3,773        & 6                & Stock Prices            & time series    & \cite{TimeGAN}          \\
                              & Monash TSF          & $30\times T$       & $V$        & Multiple                & time series    & \cite{Monash-TSF}       \\
                              & LOTSA	              & $105 \times T$    &	$V$	        & Multiple	              & time series	   & \cite{LOTSA} \\
                              & Solar               & 52,560       & 137              & Solar Power Production  & time series    & \cite{Exchange}         \\
                              & MoJoCo              & 10,000 x 100 & 14               & Control Tasks           & time series    & \cite{P19_jhin2022exit}             \\
                              & USHCN-Climate       & 386,068      & 5                & Climatological Data     & time series    & \cite{P39_schirmer2022modeling}              \\
\midrule
\multirow{28}{*}{Classification \& Clustering} &
  UCR &
  $128 \times T$ &
  1 &
  Multiple &
  time series &
  \cite{ucr-benchmark} \\
                              & UEA                 & $30 \times T$       & $V$                & Multiple                & time series    & \cite{uea-benchmark}    \\
                              & PhysioNet Sepsis    & $40,336 \times T$   & 34               & Medical Data            & time series    & \cite{Sepsis}           \\
                              & PhysioNet ICU       & $12,000 \times T$   & 36               & Medical Data            & time series    & \cite{ICU}              \\
                              & PhysioNet ECG       & $12,186 \times T$   & 1                & Medical Data            & time series    & \cite{ECG}              \\
                              & HAR                 & 10,299       & 9                & Human Activity          & time series    & \cite{HAR}              \\
                              & EMG                 & 163          & 1                & Medical Data            & time series    & \cite{Physionet}        \\
                              & Epilepsy            & 11,500       & 1                & Brain Activity          & time series    & \cite{Epilepsy}         \\
                              & Waveform            & 76,567       & 2                & Medical Data            & time series    & \cite{P103_zhang2023co}             \\
                              & Gesture             & 440          & 3                & Hand Gestures           & time series    & \cite{Gesture}          \\
                              & MOD                 & 39,609       & 2                & Moving Object           & time series    & \cite{P110_liu2023focal}            \\
                              & PAMAP2              & 9,611        & 10               & Human Activity          & time series    & \cite{P110_liu2023focal}            \\
                              & Sleep-EEG           & 371,005      & 1                & Sleep Stages            & time series    & \cite{P63_dong2023simmtm}           \\
                              & RealWorld-HAR       & 12,887       & 9                & Human Activity          & time series    & \cite{P110_liu2023focal}            \\
                              & Speech Commands     & 5,630,975    & 20               & Spoken Words            & audio          & \cite{P19_jhin2022exit}             \\
                              & LRW                 & 13,050,000   & $64 \times 64$          & Lib Reading             & video          & \cite{LRW}              \\
                              & ESC50               & 10,000       & 1                & Environmental Sound     & audio          & \cite{ESC}              \\
                              & UCF101              & 333,000      & $320 \times 240$        & Human Activity          & video          & \cite{UCF101}           \\
                              & HMDB51              & $6,849 \times T$    & $V_{width} \times V_{height}$ & Human Activity          & video          & \cite{HMDB51}           \\
                              & Kinetics-400        & 7,656,125    & $V_{width} \times V_{height}$ & Human Activity          & video          & \cite{K400}             \\
                              & AD                  & 1,527,552    & 16               & Medical Data            & time series    & \cite{P100_wang2023contrast}             \\
                              & PTB                 & 18,711,000   & 15               & Medical Data            & time series    & \cite{P100_wang2023contrast}             \\
                              & TDBrain             & 3,035,136    & 33               & Brain Activity          & time series    & \cite{P100_wang2023contrast}             \\
 &
  MMAct &
  \begin{tabular}[c]{@{}c@{}}36,764 \\ (only number of instances)\end{tabular} &
  N/A &
  Human Activity &
  multi-modality &
  \cite{MMAct} \\
                              & PennAction          & $2,326 \times T$    & $640 \times 480$        & Human Activity          & video          & \cite{P22_chen2022frame}              \\
                              & FineGym             & $4,883 \times T$    & $V_{width} \times V_{height}$ & Human Activity          & video          & \cite{P22_chen2022frame}              \\
                              & Pouring             & $84 \times T$       & $V_{width} \times V_{height}$ & Human Activity          & video          & \cite{P22_chen2022frame}              \\
                              & Something-Something & 2,650,164    & $84 \times 84$          & Human Activity          & video          & \cite{P20_kim2023exploring}              \\
\midrule
\multirow{8}{*}{Regression} &
  TSER Archive &
  $19 \times T$ &
  $V$ &
  Multiple &
  time series &
  \cite{tser_survey} \\
                              & Neonate             & $79 \times T$       & 18               & Neonatal EEG Recordings & time series    & \cite{Neonate}          \\
                              & IEEE SPC            & $22 \times T$       & 5                & Heart Rate Monitoring   & time series    & \cite{P99_demirel2023finding}          \\
                              & DaLia               & $15 \times T$       & 11               & Heart Rate Monitoring   & time series    & \cite{P99_demirel2023finding}          \\
                              & IHEPC               & 2,075,259    & 1                & Electricity Consumption & time series    & \cite{P91_franceschi2019unsupervised} \\
                              & AEPD                & 19,735       & 29               & Appliances Energy       & time series    & \cite{P102_zhang2023trid}         \\
                              & BMAD                & 420,768      & 6                & Air Quality             & time series    & \cite{P102_zhang2023trid}         \\
                              & SML2010             & 4,137        & 18               & Smart home              & time series    & \cite{P102_zhang2023trid}         \\
\midrule
\multirow{2}{*}{Segmentation} & TSSB                & $75 \times T$       & 1                & Multiple                & time series    & \cite{ClaSP}            \\
                              & UTSA                & $32 \times T$       & 1                & Multiple                & time series    & \cite{UTSA}            \\
\midrule
\multirow{18}{*}{Anomaly Detection} &
  FD-A &
  8,184 &
  1 &
  Mechanical System &
  time series &
  \cite{FD} \\
                              & FD-B                & 13,640       & 1                & Mechanical System       & time series    & \cite{FD}               \\
                              & KPI                 & 5,922,913    & 1                & Server Machine          & time series    & \cite{P88_yue2022ts2vec}           \\
                              & TODS                & $T$            & $V$                & Synthetic Data          & time series    & \cite{TODS}             \\
                              & SMD                 & 1,416,825    & 38               & Server Machine          & time series    & \cite{SMD}              \\
                              & ASD                 & 154,171      & 19               & Server Machine          & time series    & \cite{ASD}              \\
                              & PSM                 & 220,322      & 26               & Server Machine          & time series    & \cite{PSM}              \\
                              & MSL                 & 130,046      & 55               & Spacecraft              & time series    & \cite{NASA}             \\
                              & SMAP                & 562,800      & 25               & Spacecraft              & time series    & \cite{NASA}             \\
                              & SWaT                & 944,919      & 51               & Infrastructure          & time series    & \cite{SWaT}             \\
                              & WADI                & 1,221,372    & 103              & Infrastructure          & time series    & \cite{WADI}             \\
                              & Yahoo               & 572,966      & 1                & Multiple                & time series    & \cite{P88_yue2022ts2vec}           \\
                              & TimeSeAD            & $21 \times T$       & $V$                & Multiple                & time series    & \cite{TimeseAD}         \\
                              & TSB-UAD             & $1,980 \times T$    & 1                & Multiple                & time series    & \cite{TSB-UAD}          \\
                              & UCR-TSAD            & $250 \times T$      & 1                & Multiple                & time series    & \cite{UCR-TSAD}         \\
                              & UCFCrime            & $1,900 \times T$    & $V_{width} \times V_{height}$ & Surveillance            & video          & \cite{UCFCrime} \\
                              & Oops!               & $20,338 \times T$   & $V_{width} \times V_{height}$ & Human Activity          & video          & \cite{Oops}      \\
                              & DFDC                & $128,154 \times T$  & $256 \times 256$        & Deepfake                & video          & \cite{DFDC}             \\
\midrule
\multirow{3}{*}{Retrieval} &
  EK-100 &
  89,977 &
  $V_{width} \times V_{height}$ &
  Human Activity &
  video &
  \cite{EK-100} \\
                              & HowTo100M           & 136,600,000  & $V_{width} \times V_{height}$ & Human Activity          & video          & \cite{HowTo100M}        \\
                              & MUSIC               & 714          & $V_{width} \times V_{height}$ & Musical Instrument      & multi-modality & \cite{MUSIC}            \\ 
\bottomrule
\end{tabular}%
}
\end{table}

\subsubsection{Retrieval} Although we find that a few studies of the reviewed articles use particular datasets for the retrieval task (e.g., EK-100\,\cite{EK-100}, Howto100M\,\cite{HowTo100M}, and MUSIC\,\cite{MUSIC}), the task itself can be evaluated with any benchmark dataset because it is basically based on arbitrary query time series. For time-series retrieval\,\cite{P01_chen2021deep}, benchmark datasets for classification (e.g., UCR) are commonly used. For the evaluation, the top-$k$ recall rate (higher is better) is the standard metric to examine the overlap percentage of the top-$k$ results and the ground truth. $k$ is usually set to 5, 10, and 20.

\noindent
Finally, while this survey focuses on DL-based representation learning, we strongly encourage researchers to also benchmark their methods against traditional baselines. As recent studies\,\cite{TFB_Fair_TSF_2024, Elephant_NeurIPS_2024} have shown, these classical methods can be surprisingly competitive and provide a crucial reference for gauging the true practical benefits of more complex representation learning approaches.

\vspace*{-0.2cm}
\subsection{Additional Metric for Inherent Representation Quality}
Besides the task-specific metrics, recent studies\,\cite{P82_wu2023timesnet, P63_dong2023simmtm} also evaluate the inherent quality of the learned representation by calculating the centered kernel alignment\,(CKA) similarity between representations from the first and the last layers. A higher CKA similarity indicates more similar representations. As the bottom layer representations usually contain low-level or detailed information, a smaller similarity means the top layer contains different information from the bottom layer and indicates the model tends to learn high-level representations or more abstract information. 

In general, better forecasting and anomaly detection accuracy is related to higher CKA similarity, while better imputation and classification results corresponds to the lower CKA similarity. Specifically, the lower CKA similarity means that the representations are distinguishing among different layers, thus being hierarchical representations. These results also indicate the property of representations that each task requires. Thus, we can use this metric to evalaute whether an encoder learn appropriate representations for different tasks. 

\section{Open Challenges and Future Research Directions} \label{section:challenges_and_trends}
In this section, we discuss open challenges and outline promising future research directions that have the potential to enhance the existing universal time-series representation learning methods.

\subsection{Time-Series Active Learning}
Efforts in the DL community have aimed to address label sparsity, a critical problem for time-series data where annotation is notoriously difficult and expensive. Active learning\,(AL)\,\cite{shin2022_active, rana2022_active} has emerged as a powerful paradigm to mitigate this by intelligently selecting the most informative unlabeled instances for an expert to annotate. However, applying AL to time series presents unique challenges. First, the high dimensionality and lack of intuitive patterns can make it difficult for human annotators to provide accurate labels, necessitating effective visualization or feature extraction within the AL loop itself\,\cite{AL_TSC_2021}. Second, many applications suffer from a cold-start problem where no initially-labeled examples exist, requiring novel bootstrapping strategies like unsupervised pre-clustering. Finally, designing effective query strategies for temporally correlated data is more complex than for i.i.d. data\,\cite{AL_TS_2005}. Developing robust AL frameworks tailored to these challenges is a critical research direction.

\subsection{Distribution Shifts and Adaptation}
When a time series model undergoes continuous testing, especially during the model deployment, various forms of concept drift, such as sudden, gradual, incremental, and recurrent drifts, may occur due to the accumulation of unseen data\,\cite{agrahari2022concept}. A domain shift problem may also arise due to differences between the source domain used in the training phase and the target domain used in the test phase. Given that distribution shifts resulting from concept drift and domain shift are factors that degrade model performance, previous studies focus on concept drift adaptation and domain adaptation to address these shifts in specific downstream tasks, such as classification and forecasting\,\cite{yuan2022recent, ragab2023adatime, jin2022domain, CLUDA_ICLR}. \citet{OOD_TS_Survey_26} introduce a dedicated taxonomy of OOD generalization in time series. Our discussion complements that OOD-centric perspective by emphasizing how universal representations can remain transferable under such shifts. Promising directions include learning invariant temporal factors, separating stable components from environment-specific components, and adapting representations with discrepancy-based, adversarial, test-time, or uncertainty-aware objectives. These directions are particularly important under frozen-encoder evaluation, where a representation must remain useful across downstream tasks without extensive retraining.

\subsection{Reliable Data Augmentation}
As minor data augmentation can significantly impact time-series properties, determining the appropriate type and degree of augmentation is essential. Various techniques, including jittering, shifting, and warping, have been used for time-series representation learning. However, the reliability of these methods has not been fully explored. Data augmentation becomes more important with new learning paradigms like contrastive learning. It serves not only to expand the size of datasets but also to provide diverse class-invariant features. Unfortunately, current approaches focus more on the evolving role of data augmentation, often overlooking its fundamental task to maintain the integrity of the original data characteristics. Recent studies\,\cite{P76_luo2023time, P99_demirel2023finding} have focused on improving data augmentation reliability using innovative approaches and adaptive strategies based on more reliable criteria. Many studies still use empirical approaches to determine data augmentation, despite these advances. Therefore, methods for time-series data augmentation that estimates the reliability and efficacy of selecting the optimal augmentation strategy is promising.

\subsection{Neural Architecture Search (NAS)}
Making efficient deep models laboriously relies on tedious manual trial-and-error to design the network architectures and select corresponding hyperparameters relevant to the three design elements we discussed. Accordingly, recent endeavors employ NAS to discover an optimal architecture by automatically designing neural architectures and their hyperparameters, e.g., \# layers and network types. These configurations significantly affect the representation quality and the downstream performance. Even though NAS has demonstrated its successes in diverse tasks\,\cite{NAS_1000_Insights}, NAS for time series is still underexplored. There are only NAS methods for a specific time-series analysis, such as forecasting and classification, having limited generalizability\,\cite{LightCTS_SIGMOD_2023, AutoTransformer, trirat2024pasta, TFAS_MLJ_2025}. Therefore, NAS for universal time-series representation learning that perform well across downstream tasks is an important future direction, especially for industry-scale time series with high dimensions and large volumes newly generated every day.

\subsection{Multi-Modal Representation Learning}
Vision-language models, e.g., CLIP\,\cite{CLIP} and ALIGN\,\cite{jia2021scaling}, have recently shown remarkable performance in zero-shot learning and fine-tuning for various vision-related tasks by using the semantics of human languages. If time-series data entail semantics understandable to humans, we can make multi-view representations of time series by using human languages as an additional modality to annotate them. As a result, the learned representations from time series-language models will become more expressive and fine-grained in semantics. For example, human activity data or time-series anomalies can be described in human languages based on human movement and the context of abnormal situations. However, unlike video representation learning, which can leverage a vision-language pre-trained model for the annotation, time-series data annotation with human language can be laborious and highly domain-specific. Thus, building a large multi-modal time series-text dataset is a promising research direction.

\subsection{Interpretability, Fairness, and Responsible Use}
As universal time-series representations are increasingly used in high-stakes domains such as healthcare, finance, infrastructure, and industrial monitoring, their decisions should be interpretable, auditable, and fair across populations, sites, and sensor configurations. A key risk is that self-supervised or pre-trained representations may encode demographic, institution-specific, device-specific, or environment-specific biases that remain hidden during downstream fine-tuning. Future work should therefore develop attribution and auditing tools for universal representations, evaluate whether representation spaces preserve or amplify sensitive group differences, and design fairness-aware objectives or adaptation procedures. Shapelet- and motif-based methods and LLM-aligned representations may provide useful starting points because they can connect latent features to human-interpretable temporal patterns or textual explanations.
\section{Conclusions} \label{section:conclusion}
This article introduces \emph{universal} time-series representation learning research and its importance for downstream time-series analysis. We present a comprehensive and up-to-date literature review of universal representation learning for time series by categorizing the recent advancements from design perspectives. Our main goal is to answer how each fundamental design element---\emph{training data}, \emph{neural architectures}, and \emph{learning objectives}---of state-of-the-art methods contributes to the improvement of the learned representation quality, resulting in a novel structured taxonomy with \textbf{26} subcategories. Although most studies consider all design elements in their methods, only one or two elements are newly proposed. Based on the selected studies, we find that decomposition, transformation, and sample selection methods in the data-centric approaches are still underexplored. In addition, we provide a practical guideline about standard experimental setups and widely used time-series datasets, together with discussions on various promising research directions. Ultimately, we expect this survey to be a valuable resource for practitioners and researchers interested in a multi-faceted understanding of the universal representation learning methods for time series.

\begin{acks}
This work was supported by Institute of Information \& Communications Technology Planning \& Evaluation\,(IITP) grant funded by the Korea government\,(MSIT) (No.\ RS-2020-II200862, DB4DL: High-Usability and Performance In-Memory Distributed DBMS for Deep Learning, 50\% and No.\ RS-2022-II220157, Robust, Fair, Extensible Data-Centric Continual Learning, 50\%).
\end{acks}

\bibliographystyle{ACM-Reference-Format}
\bibliography{9-References}

\end{document}